\documentclass[10pt,twocolumn,letterpaper]{article}

\usepackage[pagenumbers]{cvpr}  
\usepackage{amsfonts}  
\usepackage{boldline}
\usepackage{subcaption}
\usepackage{graphicx}
\usepackage{adjustbox}
\usepackage{makecell}
\usepackage{amssymb}
\usepackage{pifont}
\newcommand{\cmark}{\ding{51}}%
\newcommand{\xmark}{\ding{55}}%

\newcommand\blfootnote[1]{ 
  \begingroup
  \renewcommand\thefootnote{}\footnote{#1} %
  \addtocounter{footnote}{-1} %
  \endgroup
}

\usepackage[dvipsnames]{xcolor}

\definecolor{cvprblue}{rgb}{0.21,0.49,0.74}
\usepackage[pagebackref,breaklinks,colorlinks,citecolor=cvprblue]{hyperref}

\hypersetup{
    colorlinks=true,
    urlcolor=magenta, 
}

\newcommand{\algname}{IMPRINT}

\def\paperID{7884} 
\def\confName{CVPR}
\def\confYear{2024}

\title{IMPRINT: Generative Object Compositing by Learning Identity-Preserving Representation}

\author{Yizhi Song$^{1*}$, Zhifei Zhang$^2$, Zhe Lin$^2$, Scott Cohen$^2$, Brian Price$^2$, \\
Jianming Zhang$^2$, Soo Ye Kim$^2$, He Zhang$^2$, Wei Xiong$^2$, Daniel Aliaga$^1$ \\
Purdue University$^1$, Adobe Research$^2$
}

\begin{document}

\twocolumn[{
\renewcommand\twocolumn[1][]{#1}
\maketitle
\vspace{-25pt}
\begin{center}
    \includegraphics[width=1.0\linewidth]{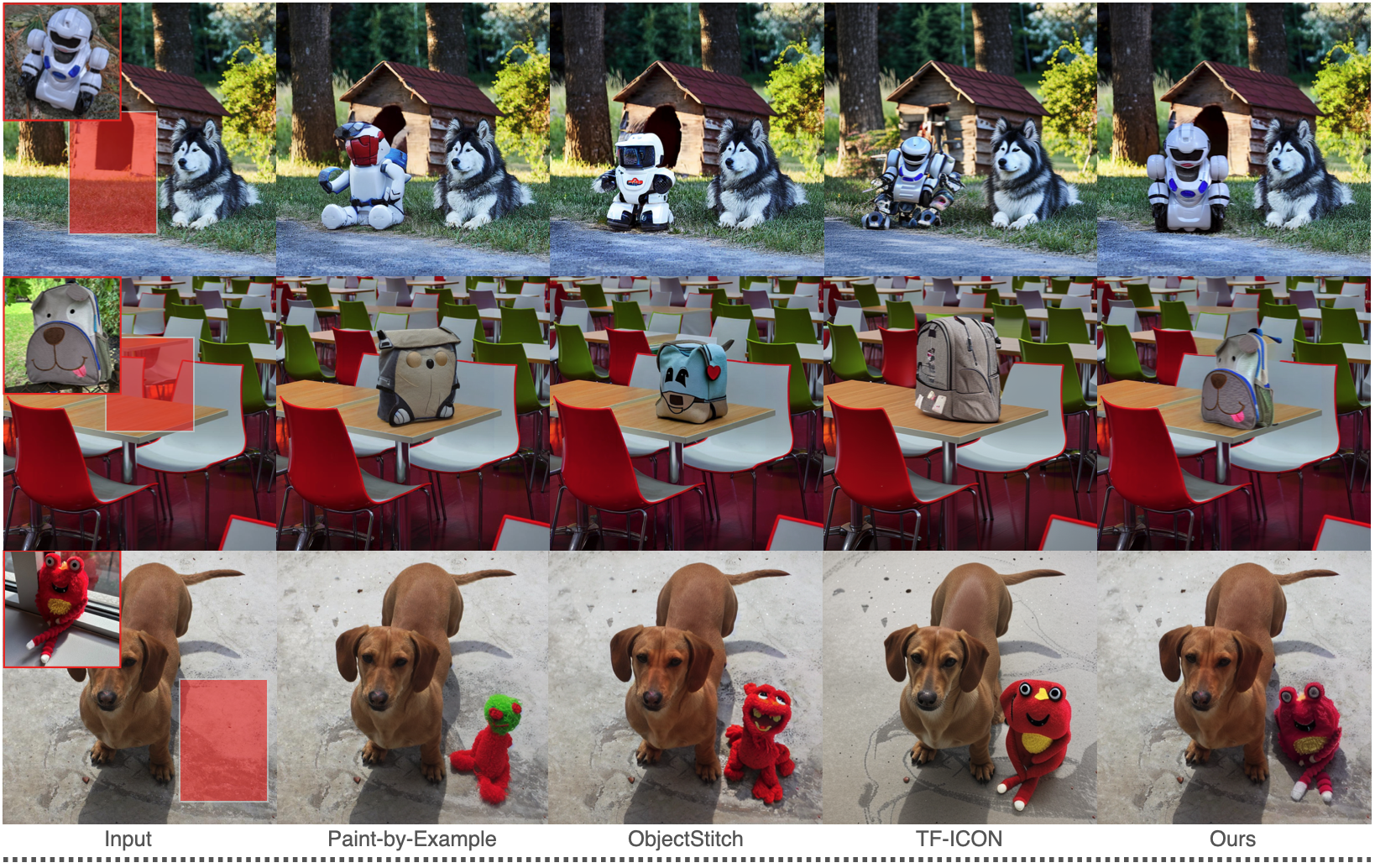}
    \includegraphics[width=1.0\linewidth]{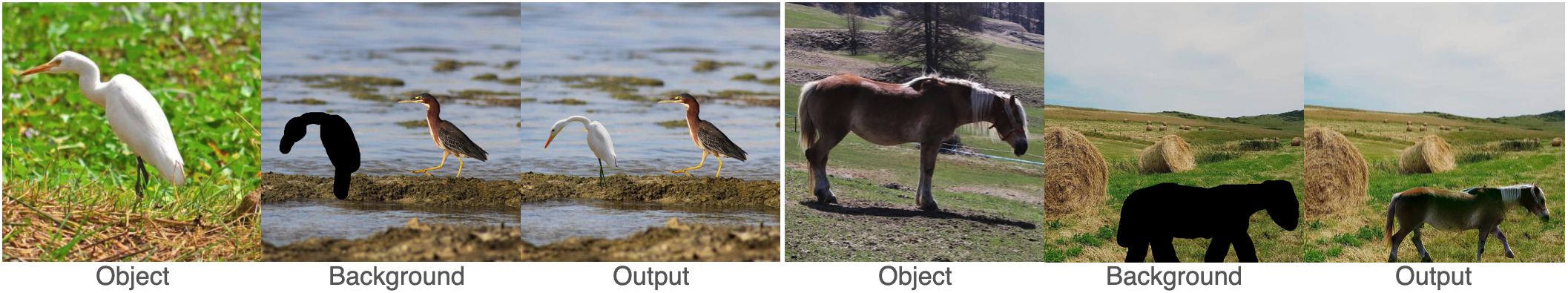}
    \captionof{figure}{Top: Comparison with three prior works, \ie, Paint-by-Example~\cite{yang2023paint}, ObjectStitch~\cite{song2023objectstitch}, and TF-ICON~\cite{lu2023tf}. Our method \textbf{\algname} outperforms others in terms of identity preservation and color/geometry harmonization. Bottom: Given a coarse mask, \textbf{\algname} can change the pose of the object to follow the shape of the mask.}
    \label{fig:teaser}
\end{center}
}]


\blfootnote{* Work done during an internship at Adobe.}


\clearpage
\begin{abstract}

Generative object compositing emerges as a promising new avenue for compositional image editing. However, the requirement of object identity preservation poses a significant challenge, limiting practical usage of most existing methods. In response, this paper introduces \textbf{\algname}, a novel diffusion-based generative model trained with a two-stage learning framework that decouples learning of identity preservation from that of compositing. 
The first stage is targeted for context-agnostic, identity-preserving pretraining of the object encoder, enabling the encoder to learn an embedding that is both view-invariant and conducive to enhanced detail preservation. The subsequent stage leverages this representation to learn seamless harmonization of the object composited to the background.
In addition, \algname{} incorporates a shape-guidance mechanism offering user-directed control over the compositing process.
Extensive experiments demonstrate that \algname{} significantly outperforms existing methods and various baselines on identity preservation and composition quality.
Project page: \href{https://song630.github.io/IMPRINT-Project-Page/}{\textnormal{https://song630.github.io/IMPRINT-Project-Page/}}

\end{abstract}
    
\section{Introduction}
\label{sec:intro}

Image compositing, the art of merging a reference object with a background to create a cohesive and realistic image, has witnessed transformative advancements with the advent of diffusion models (DM) \cite{ho2020denoising, rombach2022high, saharia2022photorealistic, ramesh2022hierarchical}. These models have catalyzed the emergence of generative object compositing, a novel task that hinges on two critical aspects: identity (ID) preservation and background harmonization. The goal is to ensure that the object in the composite image retains its identity while adapting its color and geometry for seamless integration with the background. Existing methods \cite{yang2023paint, song2023objectstitch, lu2023tf} demonstrate impressive capabilities in generative compositing; however, they often fail in ID-preservation or context consistency.

Recent works \cite{yang2023paint, song2023objectstitch}, typically struggle with balancing ID preservation and background harmony. While these methods have made strides in spatial adjustments, they predominantly capture categorical rather than detailed information. TF-ICON \cite{lu2023tf} and two concurrent works \cite{chen2023anydoor, zhang2023controlcom}  have advanced subject fidelity but at the expense of limiting pose and view variations for background integration, thus curtailing their applicability in real-world settings.

To address the trade-off between identity preservation with pose adjustment for background alignment, we introduce \algname, a novel two-stage compositing framework that excels in ID preservation. Diverging from previous works, \algname{} decouples the compositing process into ID preservation and background alignment stages. The first stage involves a novel context-agnostic ID-preserving training, wherein an image encoder is trained to learn view-invariant features, crucial for detail engraving. The second stage focuses on harmonizing the object with the background, utilizing the robust ID-preserving representation from the first stage. This bifurcation allows for unprecedented fidelity in object detail while facilitating adaptable color and geometry harmonization.

Our contributions can be summarized as follows:
\begin{itemize}
    \item We introduce a novel context-agnostic ID-preserving training, demonstrating superior appearance preservation through comprehensive experiments.
    \item Our two-stage framework distinctively separates the tasks of ID preservation and background alignment, enabling realistic compositing effects.
    \item We incorporate mask control into our model, enhancing shape guidance and generation flexibility.
    \item We conduct an extensive study on appearance retention, offering insights into various factors influencing identity preservation, \eg, image encoders, multi-view datasets, training strategies, etc.
\end{itemize}

\section{Related Work}
\label{sec:related_work}

\subsection{Image Compositing}

Image compositing, a pivotal task in image editing applications, aims to insert a foreground object into a background image seamlessly, striving for realism and high fidelity.

Traditionally, image harmonization \cite{jiang2021ssh, xue2022dccf, guerreiro2023pct, ke2022harmonizer} and image blending \cite{perez2003poisson, zhang2020deep, zhang2021deep, wu2019gp} focus on color and lighting consistency between the object and the background.
However, these approaches fall short in addressing geometric adjustments.
GAN-based works \cite{lin2018st, chen2019toward, azadi2020compositional} target geometry inconsistency, yet are often domain-specific (\eg, indoor scene) and limited in handling complex transformations (\eg, out-of-plane rotation).
Shadow synthesis methods like SGRNet \cite{hong2022shadow} and PixHt-Lab \cite{sheng2023pixht} focus on realistic lighting effects.

With the advent of diffusion models \cite{ho2020denoising, sohl2015deep, song2019generative, rombach2022high}, recent research has shifted towards unified frameworks encompassing all aspects of image compositing. Methods like \cite{yang2023paint, song2023objectstitch} employ CLIP-based adapters for leveraging pretrained models, but they struggle in preserving the object's identity due to their focus on high-level semantic representations. While TF-ICON \cite{lu2023tf} improves fidelity by incorporating noise modeling and composite self-attention injection, it faces limitations in object pose adaptability.

Recent research is increasingly centering on appearance preservation in generative object compositing. Two concurrent works, AnyDoor~\cite{chen2023anydoor} and ControlCom~\cite{zhang2023controlcom}, have made strides in this area. AnyDoor combines DINOv2 \cite{oquab2023dinov2} and high-frequency filter, and ControlCom introduces a local enhancement module. However, these models have limited spatial correction capabilities. 
 In contrast, our model designs a novel approach that substantially enhances visual consistency of the object while maintaining geometry and color harmonization, representing a significant advancement in the field.


\subsection{Subject-Driven Image Generation}

Subject-driven image generation, the task of creating a subject within a novel context, often involves customizing subject attributes based on text prompts. Based on diffusion models, \cite{gal2022textual, kawar2023imagic} have led to techniques like using placeholder words for object representation, enabling high-fidelity customizations. Subsequent works \cite{ruiz2023dreambooth, ruiz2023hyperdreambooth, kumari2023multi, liu2023cones} extend this by fine-tuning pretrained text-to-image models for new concept learning. These advancements have facilitated diverse applications, such as subject swapping \cite{gu2023photoswap}, open-world generation \cite{li2023gligen}, and non-rigid image editing \cite{cao2023masactrl}. However, these methods usually require inference-time fine-tuning or multiple subject images, limiting their practicality.
In contrast, our framework offers a fast-forward and background-preserving approach that is versatile for a broad spectrum of real-world data.

\begin{figure*}
\centering
\begin{subfigure}[][][t]{0.49\textwidth}
    \includegraphics[width=\textwidth]{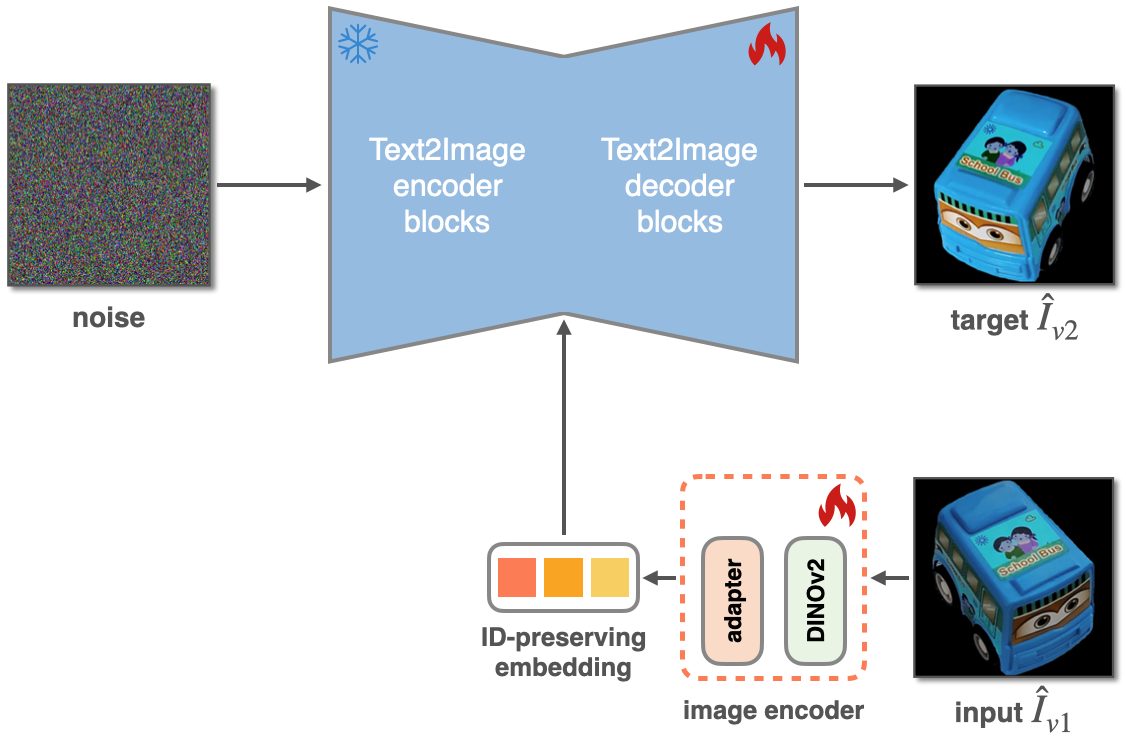}
    \caption{Stage of context-agnostic ID-preserving: we design a novel image encoder (with pre-trained DINOv2 as backbone) trained on multi-view object pairs to learn view-invariant ID-preserving representation.}
    \label{fig:stage1}
\end{subfigure}
\hspace{1pt}
\unskip \vrule 
\hspace{1pt}
\begin{subfigure}[][][t]{0.49\textwidth}
    \includegraphics[width=\textwidth]{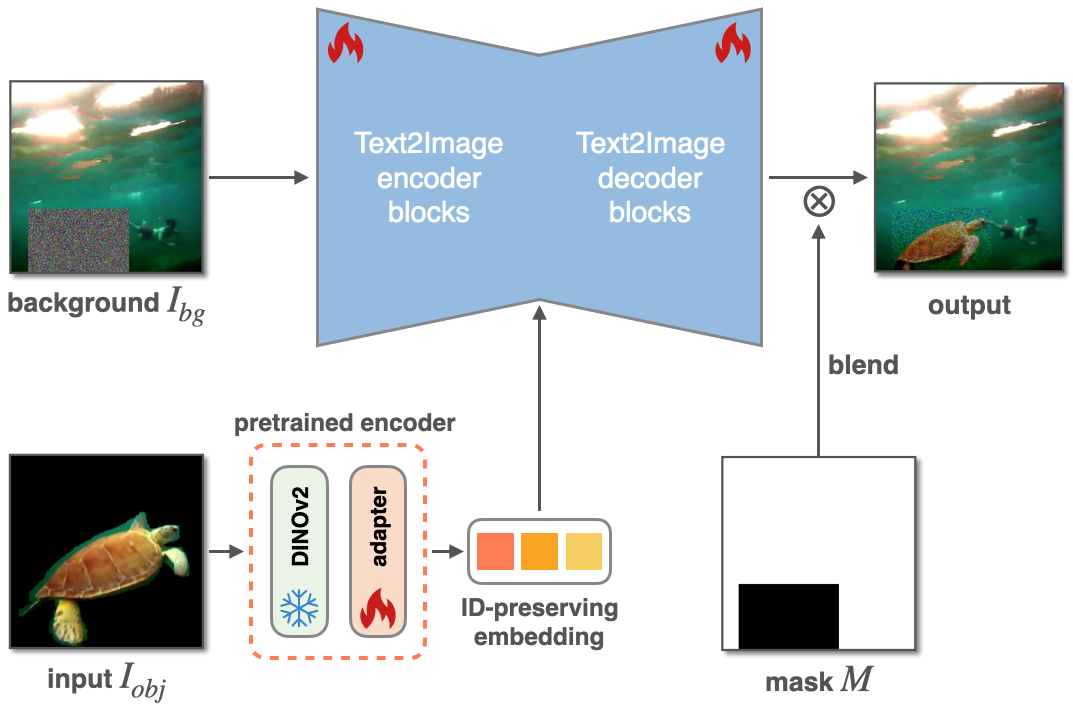}
    \caption{Stage of object compositing: taking the learned image encoder from the first stage and freezing its backbone, the whole model is trained for compositing the object to the masked region (see \cref{fig:blend} for the blending process).}
    \label{fig:stage2}
\end{subfigure}
\hfill
\caption{The two-stage training pipeline of the proposed \algname.}
\label{fig:main_pipeline}
\end{figure*}

\section{Approach}
\label{method}

\begin{figure}[t]
  \centering
   \includegraphics[width=1.0\linewidth]{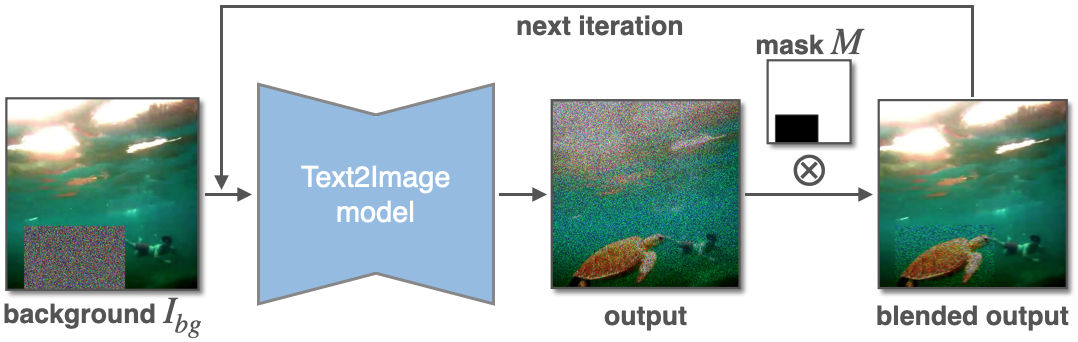}
   \vspace{-10pt}
   \caption{Illustration of the background-blending process. At each denoising step, the background area of the denoised latent is masked and blended with unmasked area from the clean background (intuitively, the model is only denoising the foreground).}
   \label{fig:blend}
\end{figure}


The proposed object compositing framework, \algname{}, is summarized in \cref{fig:main_pipeline}. Formally, given input images of object $I_{obj} \in \mathbb{R}^{H \times W \times 3}$, background $I_{bg} \in \mathbb{R}^{H \times W \times 3}$, and mask $M \in \mathbb{R}^{H \times W}$ that indicates the location and scale for object compositing to the background, we aim to learn a compositing model $\mathcal{C}$ to achieve a composite image $I_{out} = \mathcal{C} (I_{obj}, I_{bg}, M) \in \mathbb{R}^{H \times W \times 3}$. 
The ideal outcome is an $I_{out}$ that appears visually coherent and natural, \ie,  $\mathcal{C}$ should ensure that the composited object retains the identity of $I_{obj}$, aligns to the geometry of $I_{bg}$, and blends seamlessly into the background.

In this section, we expand upon our approach. To leverage pretrained text-to-image diffusion models, we design a novel image encoder to replace the text-encoding branch, thus retaining much richer information from the reference object (see \cref{method_stage1}). Distinct from existing works, our pipeline bifurcates the task into two specialized sub-tasks to concurrently ensure object fidelity and allow for geometric variations. The first stage defines a context-agnostic ID-preserving task, where the image encoder is trained to learn a unified representation of generic objects (\cref{method_stage1}). The second stage mainly trains the generator for an image compositing task (\cref{method_stage2}). In addition, we delve into various aspects contributing to the detail retention capability of our framework: \cref{method_data} discusses the process of paired data collection, and \cref{method_training} details our training strategy.

\subsection{Context-Agnostic ID-preserving Stage}
\label{method_stage1}

Distinct from prior methods, we introduce a supervised object view reconstruction task as the first stage of the training that help identity preservation. The motivation behind this task is based on the following key observations:

\begin{itemize}
    \item Existing efforts \cite{chen2023anydoor, zhang2023controlcom, lu2023tf}, which successfully improve detail preservation, are limited in geometry harmonization and tend to demonstrate copy-and-paste behavior.
    \item There is a fundamental trade-off between identity preservation and image compositing: the object is expected to be altered, in terms of color, lighting, and geometry, to better align with the background, while simultaneously, the object's original pose, color tone, and illumination effects are memorized by the model and define its appearance.
    \item Multi-view data plays a significant role in keeping identity, yet acquiring such datasets is costly. Most large-scale multi-view datasets (\cite{deitke2023objaverse, yu2023mvimgnet}) lack sufficient contextual information for compositing; they either lack a background entirely or have a background area that is too limited.
\end{itemize}

Based on the above insights, we give a formal definition of the task (as depicted in \cref{fig:stage1}): given an object of two views $I_{v1}, I_{v2}$ and their associated masks $M_{v1}, M_{v2}$, the background is removed and the segmented object pairs are denoted as $\hat{I}_{v1} = I_{v1} \bigotimes M_{v1}$, $\hat{I}_{v2} = I_{v2} \bigotimes M_{v2}$. We build a view synthesis model $\mathcal{S}=\{\mathcal{E}_u, \mathcal{G}_{\theta}\}$ conditioned on $\hat{I}_{v1}$ to generate the target view $\hat{I}_{v2}$, where $\mathcal{E}_u$ is the image encoder and $\mathcal{G}_{\theta}$ is the UNet backbone parameterized by $\theta$. 

\textbf{Image Encoder}
$\mathcal{E}_u$ consists of a pretrained DINOv2 \cite{oquab2023dinov2} and a content adapter following \cite{song2023objectstitch}. DINOv2 is a SOTA ViT model outperforming its predecessors \cite{radford2021learning, ilharco2openclip, singh2022revisiting} which extracts highly expressive visual features for reference-based generation.
The content adapter allows the utilization of pretrained T2I models by bridging the domain gap between image and text embedding spaces.

\textbf{Image Decoder} $\mathcal{G}_{\theta}$ takes the conditional denoising autoencoder $\mathcal{G}_{\theta}$ from Stable Diffusion \cite{rombach2022high} and fine-tune its decoder during training.
The objective function is defined as (based on \cite{rombach2022high}):
\begin{equation}
    \mathcal{L}_{\mbox{id}} = \mathbb{E}_{\hat{I}_{v1}, \hat{I}_{v2}, t, \epsilon} \left[ \left\| \epsilon - \mathcal{G}_{\theta}\left(\hat{I}_{v2}, t, \mathcal{E}_u\left(\hat{I}_{v1}\right)\right) \right\|_{2}^{2} \right],
    \label{eq:stage1_loss}
\end{equation}
where $\mathcal{L}_{\mbox{id}}$ is the ID-preserving loss and $\epsilon \sim \mathcal{N}(0, 1)$. The image encoder $\mathcal{E}_u$ and the decoder blocks of $\mathcal{G}_{\theta}$ are optimized in this process.
Intuitively, the encoder trained for this task will always extract representations that are view-invariant while keeping identity-related details that are shared across different views. The qualitative results of this stage are shown in \cref{exp_ablation}.
Unlike previous view-synthesis works \cite{liu2023zero},  our context-agnostic ID-preserving stage does not require any 3D information (\eg, camera parameters) as conditions, and we mainly focus on ID-preservation instead of geometrical consistency to background (which will be handled in the second stage). Therefore, only the image encoder will be taken to the next stage.

\subsection{Compositing Stage}
\label{method_stage2}

\cref{fig:stage2} illustrates the pipeline of the second stage which is trained for the compositing task, comprising the finetuned image encoder $\mathcal{E}_u$ and a generator $\mathcal{G}_{\phi}$ (parameterized by $\phi$) conditioned on the ID-preserving representations.

A simple approach is to ignore the view synthesis stage, training the encoder and generator jointly in a single-stage framework. Unfortunately, we found quality degradation from two aspects in this naive endeavor (see \cref{exp_ablation}):

\begin{itemize}
    \item When DINOv2 is trained in this stage, the model exhibits more frequent copy-paste-like behavior that composites the object in a very similar view as its original view. 
    \item When object-centric multi-view datasets, \eg, MVImgNet~\cite{yu2023mvimgnet}, are enabled in the training set, the model tends to produce more artifacts and exhibit poorer blending results due to the absence of background information in such datasets.
\end{itemize}

\noindent To overcome the issues above, we freeze the backbone of the image encoder (\ie, DINOv2) in the second stage and carefully collect a training set (see \cref{method_data} for details).

In this stage, we also leverage a pretrained T2I model as the backbone of the generator, which uses the background $I_{bg}$, a coarse mask $M$ as inputs, and is conditioned on a ID-preserving object tokens $\hat{E}_u = \mathcal{E}_u(I_{obj})$, where $I_{obj}$ indicates a masked object image. The generation is guided by injecting object tokens into the cross attention layers of $\mathcal{G}_{\phi}$.
The coarse mask also allows the synthesis of shadows, and interactions of the object and the nearby objects.

As $\hat{E}_u$ already encompasses structured view-invariant details of the object, color and geometric adjustments are no longer limited by identity preservation efforts. This freedom allows for greater variation in compositing.

We define the objective function of this stage as:
\begin{equation}
    \small
    \mathcal{L}_{\mbox{comp}} = \mathbb{E}_{I_{obj}, I_{bg}^\ast, M, t, \epsilon} \left[ M \left\| \epsilon - \mathcal{G}_{\phi}\left(I_{bg}^\ast, t, \hat{E}_u\right) \right\|_{2}^{2} \right]
    \label{eq:stage2_loss}
\end{equation}
where $\mathcal{L}_{comp}$ is the compositing loss, $I^*_{bg}$ is the target image. $\mathcal{G}_{\phi}$ and the adapter are optimized.

\textbf{The Background-blending Process}
To ensure that the transition area between the object and the background is smooth, we adopt a background-blending strategy. This process is depicted in \cref{fig:blend}.

\textbf{Shape-guided Controllable Compositing}
could enable more practical guidance of the pose and view of the generated object by drawing a rough mask. However, most prior works \cite{song2023objectstitch, lu2023tf, chen2023anydoor} have no such control. In our proposed model, following \cite{xie2022smartbrush}, masks are defined at four levels of precision (see the Appendix), where the most coarse mask is a bounding box. Incorporating multiple levels of masks replicates real-world scenarios, where users often prefer more precise masks. Results are shown in \cref{fig:teaser}.any

\subsection{Paired Data Generation}
\label{method_data}

\label{sec:data}
\begin{figure}[t]
  \centering
   \includegraphics[width=.90\linewidth]{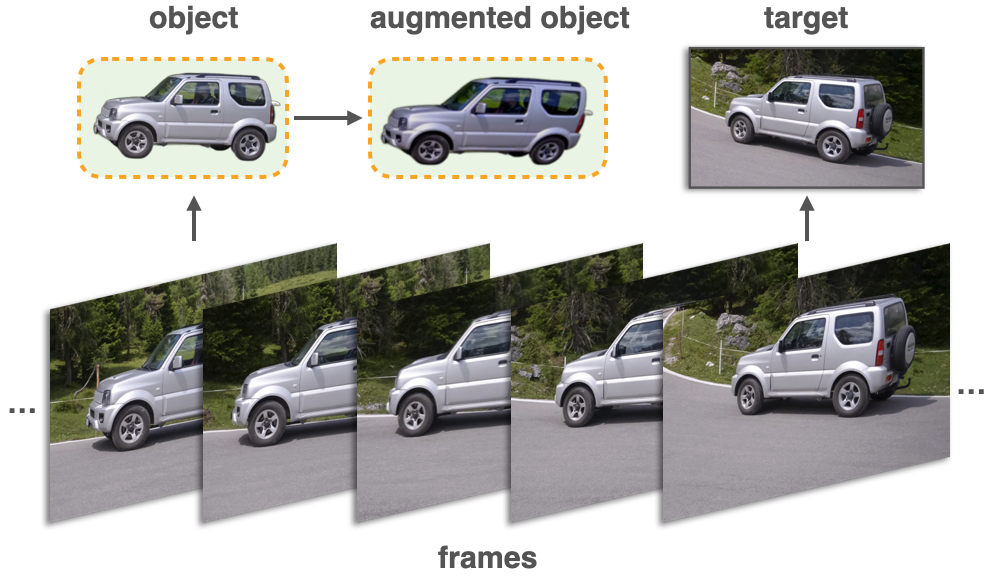}
   \caption{Illustration of the data augmentation pipeline.}
   \label{fig:data_aug}
\end{figure}

The dataset quality is another key to better identity preservation and pose variation. As proved by \cite{chen2023anydoor}, multi-view datasets can significantly improve the generation fidelity. In practice, we use a combination of image datasets (Pixabay), panoptic video segmentation datasets (YoutubeVOS \cite{xu2018youtube}, VIPSeg \cite{miao2022large} and PPR10K \cite{jie2021PPR10K}) and object-centric datasets (MVImgNet \cite{yu2023mvimgnet} and Objaverse \cite{deitke2023objaverse}). They are incorporated in different training stages and associated with various processing procedures in our self-supervised training.

The image datasets we collected have high resolution and rich background information, so they are only utilized in the second stage for better compositing. Inspired by \cite{song2023objectstitch, yang2023paint}, to simulate the lighting and geometry changes in object compositing, we design an augmentation pipeline $\hat{I}_{obj} = \mathcal{P}(\mathcal{T}(I_{obj}))$, where $\mathcal{T}$ are the affine transformations, and $\mathcal{P}$ is color and light perturbation, supported by the look-up table in \cite{jiang2021ssh}. The perturbed object $\hat{I}_{obj}$ is used as the input and the natural image $I_{bg}^\ast$ containing the original object is used as the target.

Video segmentation datasets usually suffer from low resolution and motion blur, which harm the generation quality. Nevertheless, they provide object pairs which naturally differ in lighting, geometry, view and even provide non-rigid pose variations. As a result, they are also used in the second stage. Illustrated by \cref{fig:data_aug}, each training pair comes from one video with instance-level segmentation labels. Two distinct frames are randomly sampled; one serves as the target image, while the object is extracted from the other frame as the augmented input.

Object-centric datasets offer a significantly larger scale than video segmentation datasets and provide more intricate object details. However, they are only used in the first stage due to the limited background information available in these datasets. During training, each pair $I_{v1}, I_{v2}$ are also randomly sampled from the same video with $|v1 - v2| \leq n$, where $n$ is the temporal sampling window. Empirically, we observe a loss in the generation quality as $n$ increases, and $n=7$ strikes a balance between fidelity and quality.

\subsection{Training Strategies}
\label{method_training}

All previous (or concurrent) training-free methods \cite{yang2023paint, song2023objectstitch, chen2023anydoor, zhang2023controlcom} use a \textit{frozen} transformer-based image encoder, either using DINOv2 or CLIP. However, freezing the encoder will limit their capability in extracting the object details: i) CLIP only encodes the semantic features of the object; ii) DINOv2 is trained on a dataset that is constructed based on image retrieval, allowing objects that are not entirely identical to be treated as the same instance. To overcome this challenge, we fine-tune the encoder specifically for compositing, ensuring the extraction of instance-level features.

Due to the extensive scale of the aforementioned encoders, they are prone to overfitting. The implementation of appropriate training strategies can effectively stabilize the training process and improve identity preservation. To this end, we design a novel training scheme: Sequential Collaborative Training.

More specifically, the object compositing stage is further divided into two phases: 1) in the first $n$ epochs, we assign the adapter a larger learning rate of $4\times10^{-5}$, and assign the UNet a smaller learning rate of $4\times10^{-6}$; 2) in the next $n$ epochs, we swap the learning rate of these two components (and the training finishes).
This strategy focuses on training one component at each phase, with the other component simultaneously trained at a lower rate to adapt to the changed domain; the generator is trained in the end to ensure the synthesis quality.

\section{Experiments}
\label{exp}

\subsection{Training Details}
\label{exp_details}

The first stage is trained on 1,409,545 pairs and validated on 11,175 pairs from MVImgNet, which takes 5 epochs to finish. The learning rate associated with DINOv2 (ViT-g/14 with registers) is $4 \times 10^{-6}$, and the batch size is 256. The image embedding is dropped at a rate of 0.05.

The second stage is fine-tuned on a mixture of image datasets and video datasets, including a training set of 217,451 pairs and a validation set of 15,769 pairs (listed in \cref{tab:dataset}), where we apply \cite{lee2020centermask} to obtain the segmentation masks as labels. It is trained for 15 epochs with a batch size of 256. The embedding is dropped at a rate of 0.1.

In both stages, the images are resized to $512 \times 512$. During inference, the DDIM sampler generates the composite image after 50 denoising steps using a CFG \cite{ho2022classifierfree} scale of 3.0. The model is trained on 8 NVIDIA A100 GPUs. The model is built on Stable Diffusion v1.4 (\cite{rombach2022high}).

\subsection{Evaluation Benchmark}
\label{exp_benchmark}

\textbf{Datasets} are collected from Pixabay and DreamBooth~\cite{ruiz2023dreambooth} for testing. More specifically, Pixabay testing set has 1,000 high-resolution images and has no overlap with the training set. A foreground object is selected from each image and perturbed through the data augmentation pipeline as in \cref{method_data}.
The DreamBooth testing set consists of 25 unique objects with various views. Combined with 59 background images that are manually chosen, 113 pairs are generated for this test set. This dataset is challenging since most objects are of complex texture or structure. We also conduct a user study on this dataset.

\begin{table}
  \small
  \centering
  \begin{tabular}{l|rrrr}
    \hlineB{2}
    Datasets & Pixabay & VIPSeg & YoutubeVOS & PPR10K \\ \hline
    Training   & 116,820 & 51,743 & 42,868 & 6,020 \\ 
    Validation & 6,490 & 5,487 & 3,690 & 102 \\ \hlineB{2}
  \end{tabular}
  \caption{Statistics of the datasets used in the second stage.}
  \label{tab:dataset}
\end{table}

\textbf{Metrics}
measuring fidelity and realism are adopted 
to evaluate the effectiveness of different models in terms of identity preservation and background harmonization.
We utilize CLIP-score \cite{hessel2021clipscore}, DINO-score, and DreamSim \cite{fu2023learning} as the measurements of generation fidelity. To obtain more precise comparison results, we always crop the output images so that the generated object is located in the center of the image.
FID \cite{heusel2017gans} is employed to measure the realism which indicates the compositing quality.

\subsection{Quantitative Evaluation}
\label{exp_quan}

To demonstrate the effectiveness of our model, we test our model and three baseline methods (Paint-by-Example \cite{yang2023paint}, ObjectStitch \cite{song2023objectstitch}, and TF-ICON \cite{lu2023tf}) on the two aforementioned test sets. The same inputs (a mask and a reference object) are used in all models. For fair comparison, we further fine-tune Paint-by-Example (PbE) on our second-stage training set.

When testing on TF-ICON, we employ the parameter set in "same domain" mode, as suggested by the official implementation. It also requires a text prompt as an additional input, so we apply BLIP2 \cite{li2023blip2}, a state-of-the-art vision-language model to generate captions for the images. Moreover, the captions for the DreamBooth test set are manually refined to improve the performance.
As shown in \cref{tab:quan_baselines}, IMPRINT achieves the best performance in both realism and fidelity. See the Appendix for quantitative comparisons with AnyDoor.

\begin{table}
\centering
\begin{adjustbox}{width=0.47\textwidth}
\begin{tabular}{lcccc}
\toprule
\textbf{Method} & \textbf{FID $\downarrow$} & \textbf{CLIP-score$\uparrow$} & \textbf{DINO-score$\uparrow$} & \textbf{DreamSim $\downarrow$} \\ 
\cmidrule{1-5}
PbE  & - & 71.5000 & 31.3765 & 0.4954 \\
\cmidrule{1-5}
OS   & - & 73.6250 & 32.9739 & 0.4297 \\ 
\cmidrule{1-5}
T-I  & - & 75.1250 & 39.2863 & 0.3661 \\ 
\cmidrule{1-5}
Ours & - & \textbf{77.0625} & \textbf{43.4463} & \textbf{0.2898} \\ 

\cmidrule{1-5}\morecmidrules\cmidrule{1-5}
PbE  & 23.2663 & 93.6250 & 85.2260 & 0.1907 \\
\cmidrule{1-5}
OS   & 22.4934 & 94.9375 & 90.3853 & 0.1422 \\ 
\cmidrule{1-5}
T-I  & 63.9730 & 88.3125 & 73.2155 & 0.3219 \\ 
\cmidrule{1-5}
Ours & \textbf{16.4487} & \textbf{96.1875} & \textbf{94.705} & \textbf{0.0831} \\ 
\bottomrule
\end{tabular}
\end{adjustbox}
\caption{Quantitative comparison with prior works. IMPRINT and the baselines are tested on two datasets for realism and ID-preserving measurement: DreamBooth (top) and the Pixabay test set (bottom). The results on both datasets demonstrate the advance of our model in both ID-preserving and realistic harmonization with the background.
}
\label{tab:quan_baselines}
\end{table}

\begin{table}
  \small
  \centering
  \begin{tabular}{l|ll|ll|ll}
    \hlineB{2}
     & Ours & OS & Ours & PbE & Ours & T-I \\ \hline
    Realism  & \textbf{50.68} & 49.32 & \textbf{62.84} & 37.16 & \textbf{53.38} & 46.62 \\ 
    Fidelity & \textbf{80.41} & 19.59 & \textbf{86.49} & 13.51 & \textbf{73.65} & 26.35 \\ \hlineB{2}
  \end{tabular}
  \caption{User study results. We design two questions to measure the realism and fidelity of the generation. In both questions, the user is presented side-by-side comparisons of our generated image and another image randomly chosen from one of the baselines. The results in the table show user preference percentage. Our model not only achieves better realism, but also outperforms the baselines in ID-preserving by a large margin.}
  \label{tab:user_study}
\end{table}

\subsection{Qualitative Evaluation}
\label{exp_qual}

\begin{figure*}
  \centering
  \begin{subfigure}{\linewidth}
  \includegraphics[width=\textwidth]{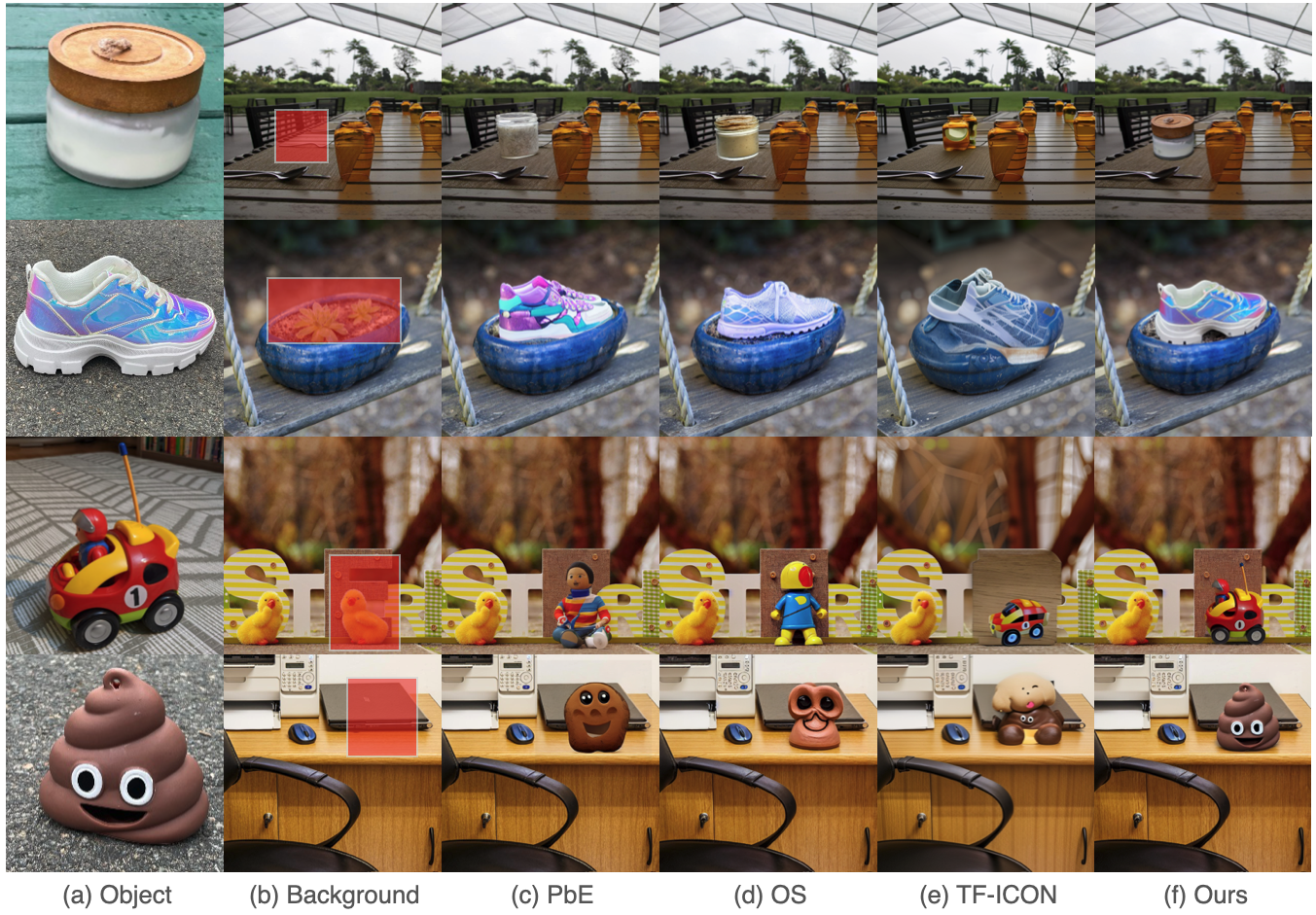}
  \end{subfigure}
  \caption{Qualitative comparison on the DreamBooth test set. Paint-by-Example and ObjectStitch lose most object details and only maintain categorical information. TF-ICON tends to copy the pose of the input subject. The comparison highlights the advantage of IMPRINT in keeping identity and making geometric changes.}
  \label{fig:qual_comp}
\end{figure*}

Qualitative comparisons are shown in \cref{fig:qual_comp}, comparing our model against prior methods. Although PbE and ObjectStitch show natural compositing effects, they often fail to capture the finer details of the objects. When the object has complex texture or structure, their generated object becomes less recognizable and even suffers from artifacts. In contrast, TF-ICON shows better consistency between the input and output, especially in keeping surface textures and captions. However, the background adaptation ability is also strictly restricted. As can be observed, TF-ICON has less variation in color and geometry changes, which results in a degradation in compositing effects. We further compare to AnyDoor in \cref{fig:anydoor} (more visual comparisons are in the Appendix). The results show that IMPRINT achieves better ID-preservation and shows the flexibility in adapting to the background in terms of color and geometry.

\begin{figure}
  \centering
  \begin{subfigure}{\linewidth}
  \includegraphics[width=\textwidth]{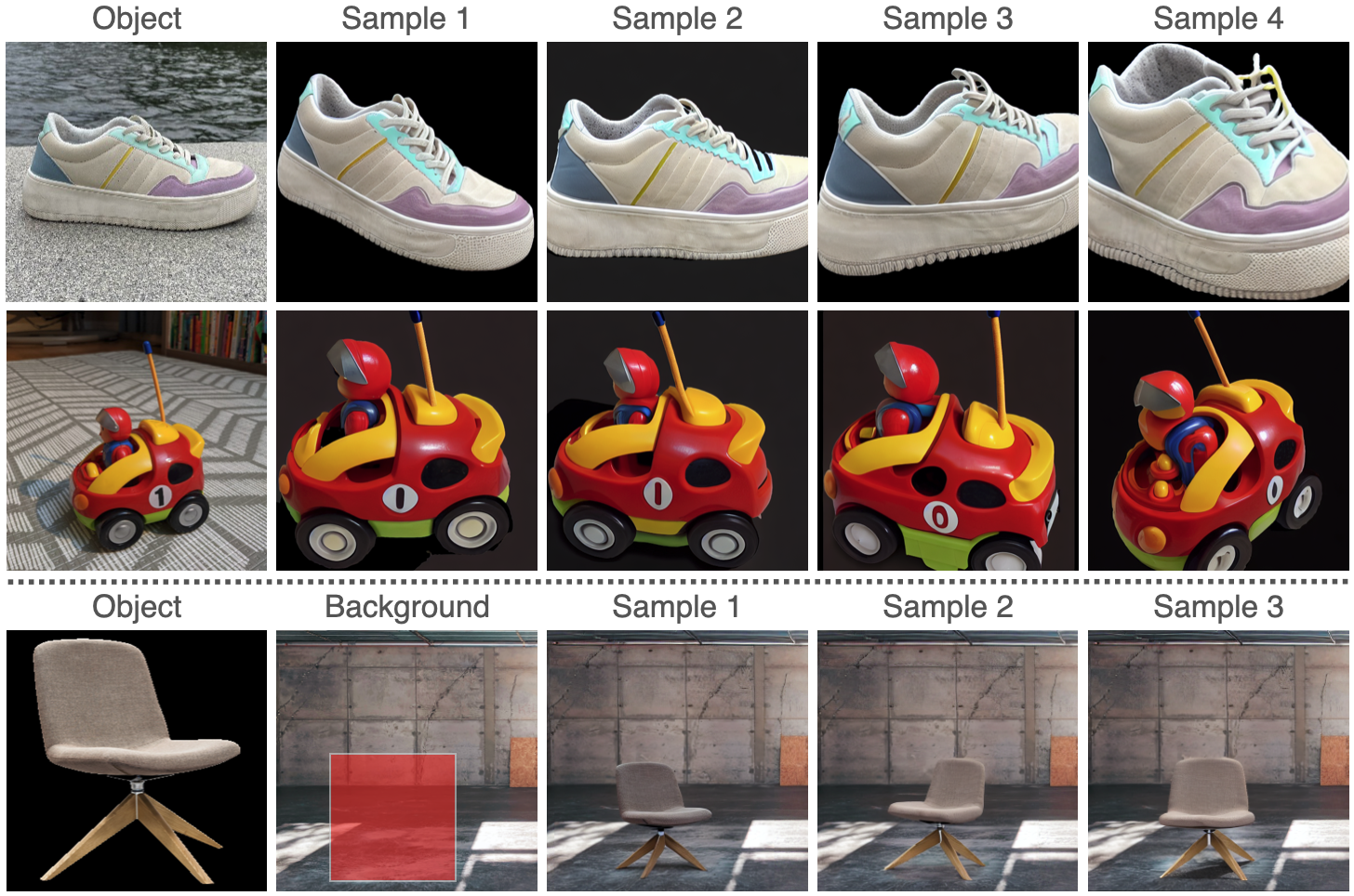}
  \end{subfigure}
  \caption{Top: Results of context-agnostic ID-preserving pretraining (after the first stage); IMPRINT generates view pose changes while memorizing the details of the object. Bottom: Diverse poses of the object after the second stage.
  }
  \label{fig:recon}
\end{figure}

We also show the synthesis results of the first stage in \cref{fig:recon}. Using the ID-preserving representation, our model is able to generate high-fidelity objects with large view variations. This process requires no extra condition such as camera parameters.

\begin{figure}
  \centering
  \begin{subfigure}{\linewidth}
  \includegraphics[width=1.0\textwidth]{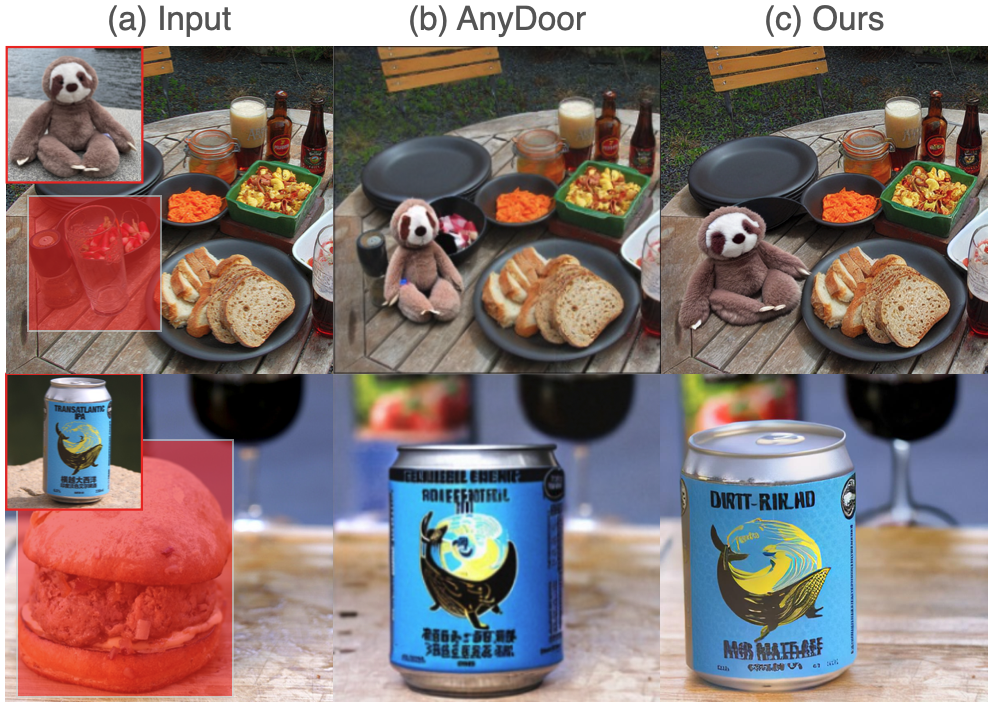}
  \end{subfigure}
  \caption{Comparison with AnyDoor \cite{chen2023anydoor}. See the Appendix for more results.
  }
  \label{fig:anydoor}
\end{figure}

\begin{figure*}
  \centering
  \begin{subfigure}{\linewidth}
  \includegraphics[width=\textwidth]{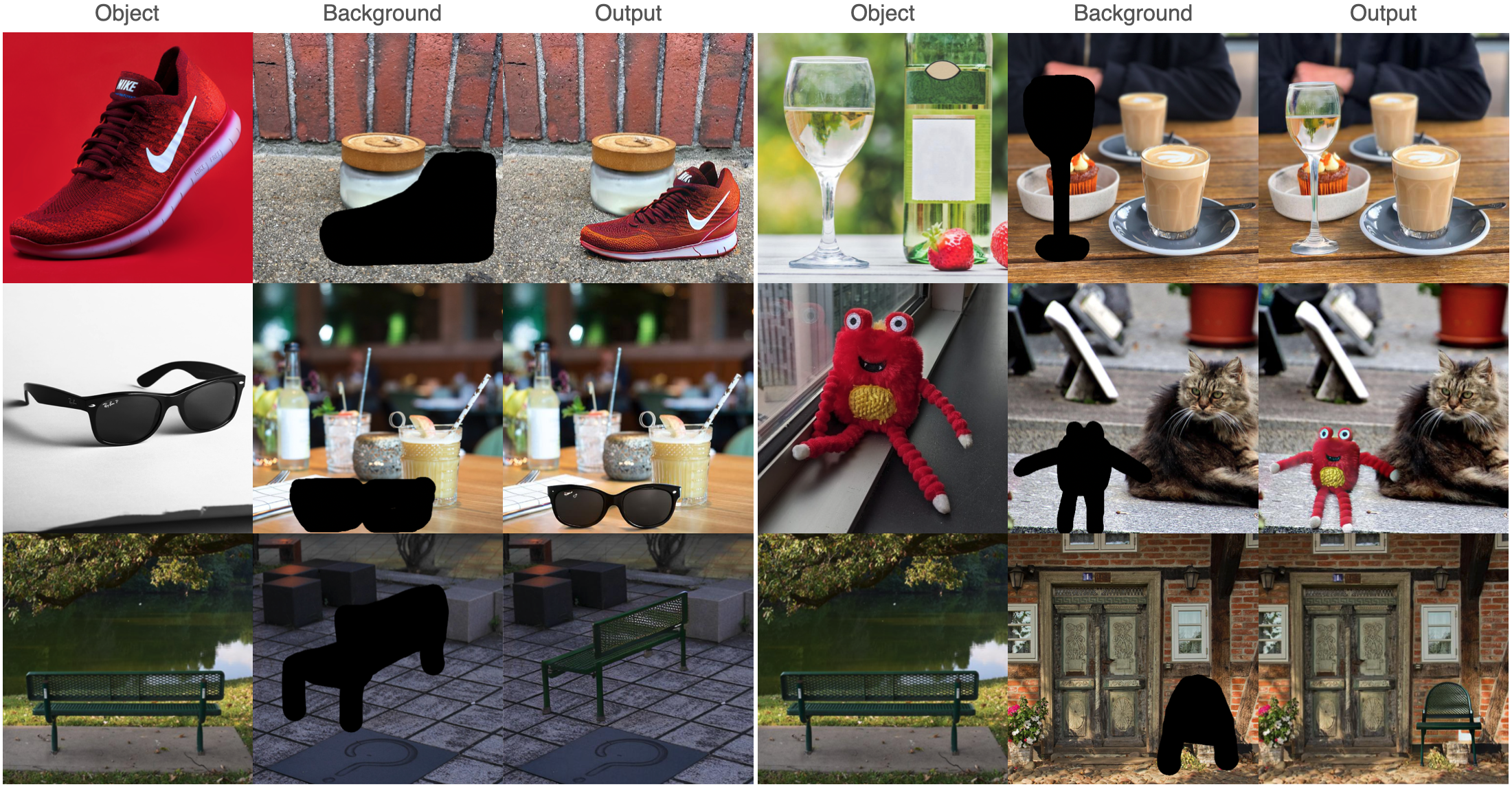}
  \end{subfigure}
  \caption{More shape-control results. IMPRINT introduces more user control by using a user-provided mask as input. Inspired by \cite{xie2022smartbrush}, we define four types of mask (including bounding box). In addition to object compositing, our model also performs edits on the input object. Depending on the shape of the coarse mask, IMPRINT can operate different types of editing, including changing the view of an object, and applying non-rigid transformation on the object.}
  \label{fig:mask_ctrl}
\end{figure*}

\begin{figure}
  \centering
  \begin{subfigure}{1.0\linewidth}
  \includegraphics[width=\textwidth]{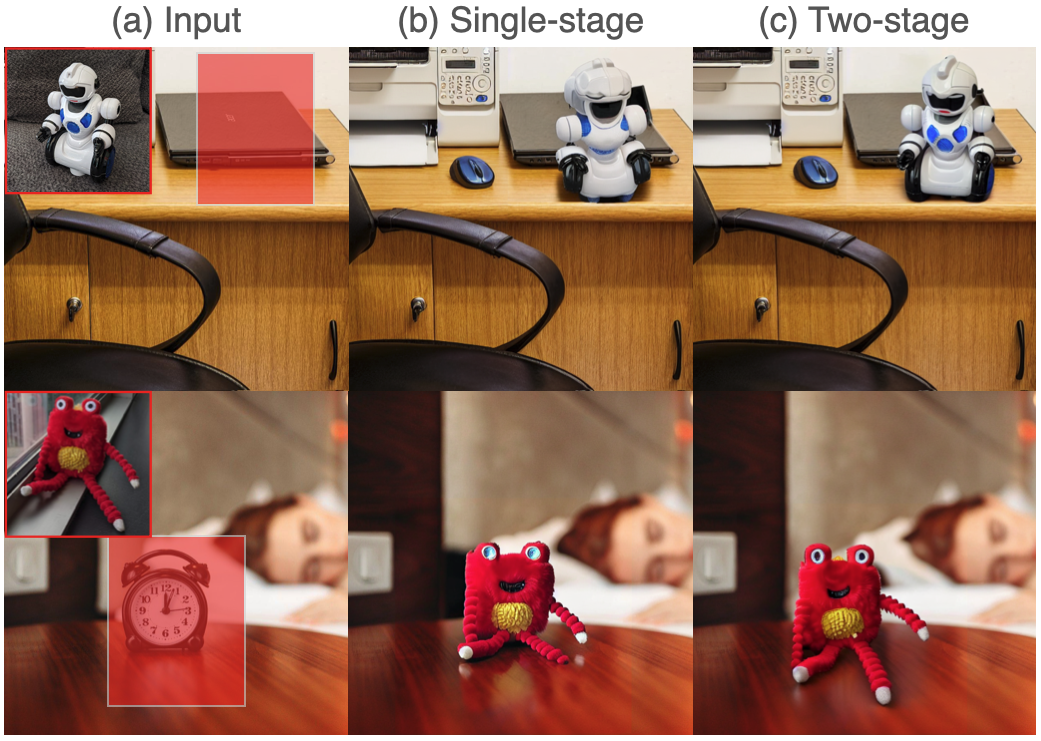}
  \end{subfigure}
  \caption{Ablation study on our two-stage training scheme. In (b) MVImgNet is added to the training set and simply trains the whole network in one stage. Compared with two-stage training, single-stage has a notable degradation in quality and loses more details.
  }
  \label{fig:ablation_pretrain}
\end{figure}

\subsection{User Study}
\label{exp_user}

We also conduct a user study using Amazon Mechanical Turk, comparing our method against the three baselines on the challenging DreamBooth dataset. The user study consists of side-by-side comparisons of our result and a randomly chosen result from the baselines. We design two questions: 1) Which image is more realistic? (the input objects are hidden from the users) 2) Which image is more similar to the reference object? Each question has 111 comparisons. We received more than 880 votes from over 130 users.
The results are shown in \cref{tab:user_study}. In terms of realism, our model outperforms PbE and TF-ICON, while comparable with ObjectStitch. We also evaluate the visual similarity. The preference rate in the table demonstrates that our method has a significant advantage over the baselines.

\begin{table*}[ht]
  \centering
  \begin{adjustbox}{width=1.0\textwidth}
  \begin{tabular}{rlccccccccc}
    \toprule
    No. & (PRE) & Encoder & Adapter & (PRE) Tune encoder & (PRE) Tune UNet & Tune encoder & Video data & MVImgNet & FID $\downarrow$ & DINO score $\uparrow$ \\
    \midrule
    1 & \xmark & CLIP   & \cmark & - & - & \xmark & \xmark & \xmark & 22.493 & 90.385 \\
    2 & \xmark & CLIP   & \cmark & - & - & \cmark & \xmark & \xmark & 19.538 & 92.695 \\
    3 & \xmark & CLIP   & \cmark & - & - & \cmark & \cmark & \xmark & 17.847 & 94.216 \\
    4 & \xmark & DINOv2 & \xmark & - & - & \xmark & \cmark & \xmark & 20.748 & 92.512 \\
    5 & \xmark & DINOv2 & \cmark & - & - & \xmark & \cmark & \xmark & 20.131 & 92.846 \\
    6 & \xmark & DINOv2 & \cmark & - & - & \cmark & \cmark & \xmark & 17.477 & 94.164 \\
    7 & \xmark & DINOv2 & \cmark & - & - & \cmark & \cmark & \cmark & 17.947 & 94.023 \\
    8 & \cmark & DINOv2 & \cmark & \cmark & \xmark & \xmark & \cmark & \xmark & 17.847 & 93.908 \\
    9 & \cmark & DINOv2 & \cmark & \xmark & \cmark & \xmark & \cmark & \xmark & 19.286 & 93.273 \\
    10 & \cmark & DINOv2 & \cmark & \cmark & \cmark & \xmark & \cmark & \xmark & \textbf{16.449} & \textbf{94.705} \\
    \bottomrule
  \end{tabular}
  \end{adjustbox}
  \caption{Ablation study on our methodologies and other common components. \textit{PRE} means whether the setting has our pretraining stage; \textit{MVImgNet} and \textit{video data} mean whether they are used in the compositing stage.
  }
  \label{tab:ablation_stage2}
\end{table*}

\subsection{Additional Visual Results of Shape-control}

Shape-guided generation introduces a lot more flexibility for image editing, as the user now gains control over the shape, view and pose of the objects, and the transformation can be either rigid or non-rigid. \cref{fig:mask_ctrl} illustrates the diverse usage of image editing given a mask as guidance.

\subsection{Ablation Study}
\label{exp_ablation}

When pursuing better identity preservation and background harmonization in the field of generative object compositing, we gain valuable experience in a wide range of techniques that contribute to this task. In \cref{tab:ablation_stage2}, we provide a complete analysis and insights of all the factors, as well as demonstrate the effectiveness of our proposed method. The same metrics are utilized as explained in \cref{exp_benchmark}.

\noindent \textbf{Training strategies.} In setting 2, we also optimize CLIP encoder. The results of settings 1 and 2 show that the optimized CLIP can capture better object identity. However, this improvement comes at the cost of variation. Setting 5 and 6 also demonstrate improved identity and less variation. For this reason, the encoder backbone is frozen in our second stage.

\noindent \textbf{Dataset.} Dataset is another component that significantly affects the performance. After adding the video datasets, the model develops a stronger capability in engraving the details (setting 2 and 3). Nevertheless, if there are too many training pairs from object-centric datasets (MVImgNet), the generation quality will degrade (setting 6 and 7) since the background information is insufficient.

\noindent \textbf{Architecture.} Inspired by \cite{song2023objectstitch}, we also use an adapter to connect the encoder with the generator. Setting 4 and 5 indicates that using the adapter will boost the overall performance in both realism and fidelity. We also observed the model converges faster when using the adapter.

\begin{figure}
\centering
\begin{subfigure}[][][c]{0.235\textwidth}
    \includegraphics[width=\textwidth]{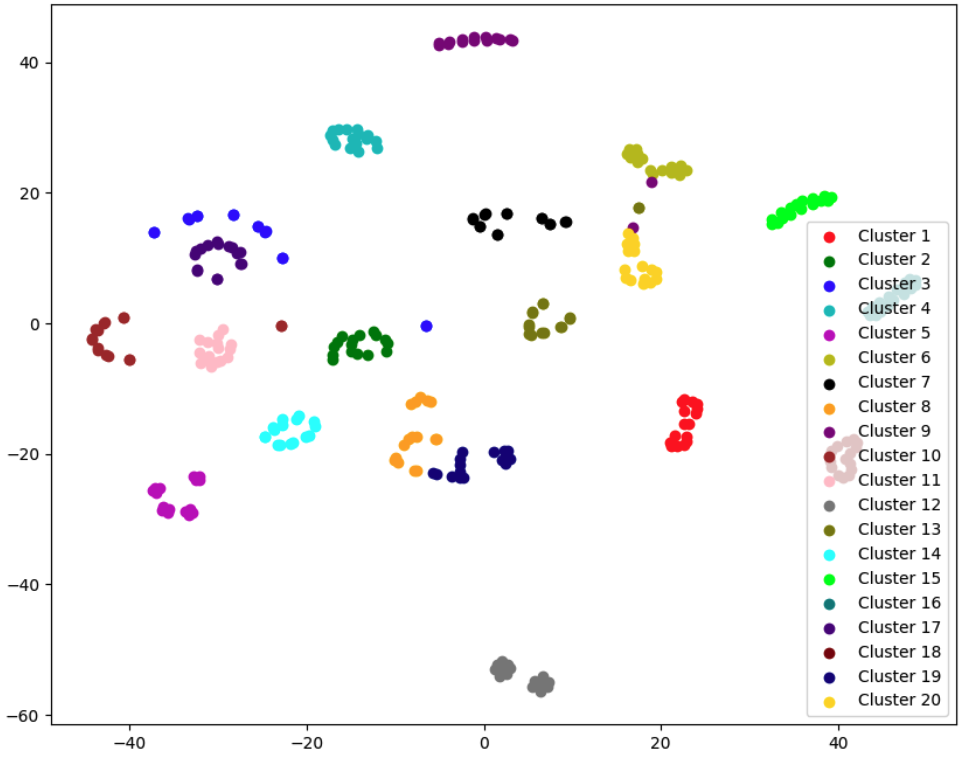}
    \caption{\centering Clustering results using the original DINOv2.}
    \label{fig:tsne1}
\end{subfigure}
\begin{subfigure}[][][c]{0.235\textwidth}
    \includegraphics[width=\textwidth]{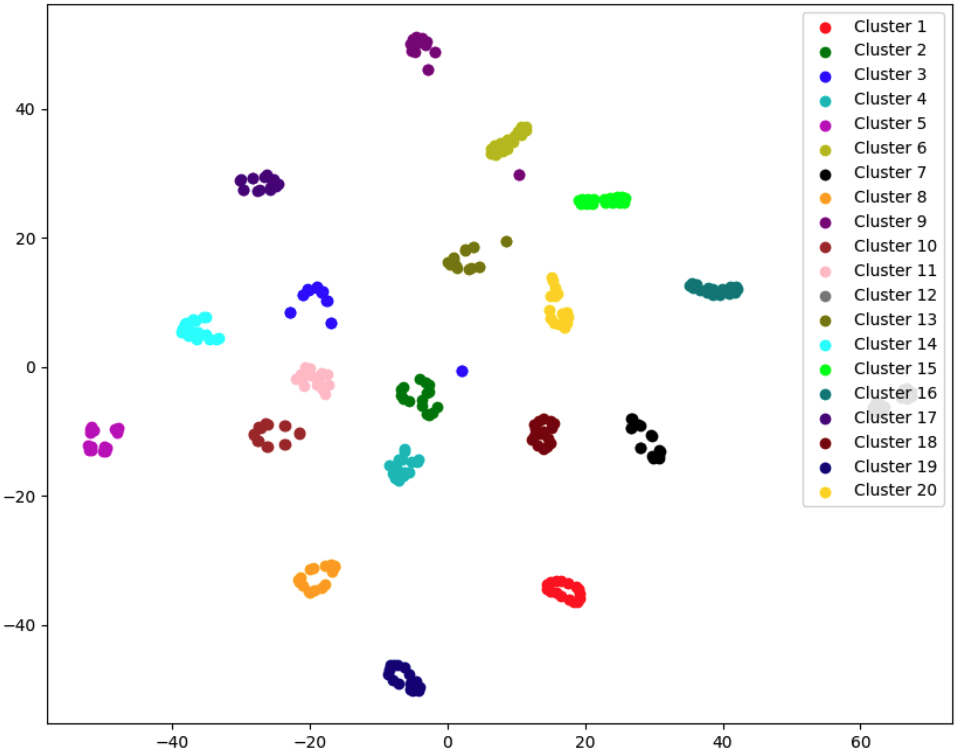}
    \caption{\centering Clustering results using our finetuned DINOv2.}
    \label{fig:tsne2}
\end{subfigure}
\hfill
\caption{Ablation study on our first stage training. We use DINOv2 (before and after the first stage) to predict embeddings of different views of 20 Objaverse objects. The embeddings are then clustered using the same algorithm and visualized using t-SNE figures. The improved clustering results demonstrate that the embeddings produced by finetuned DINOv2 have higher quality.}
\label{fig:tsne}
\end{figure}

\noindent \textbf{Pretraining.} In our framework, the first stage pretraining is a key component in improving ID-preservation and harmonization effects. To demonstrate the effectiveness, we test the original DINOv2 and our finetuned DINOv2 on a Objaverse test set. In this evaluation, the encoders generate embeddings for diverse views of 20 objects from various categories. The embeddings are then clustered and visualized in t-SNE figures (\cref{fig:tsne}). This figure shows that the finetuned encoder produces better clustering results, demonstrating that our ID-preserving representation effectively encodes the key details of the objects.
We further ablate on the first stage training using setting 7 (where there is only the compositing stage) and 10 (two-stage). Without the first stage, there is a notable drop in the compositing quality (\cref{fig:ablation_pretrain}).
Additionally, we assess the effect of freezing some components (\ie, UNet or DINOv2) during the pretraining. Compared with Setting 10, setting 8 and 9 exhibit a drop in both harmonization and ID-preservation, validating the effectiveness of our training scheme.

\section{Conclusion, Limitation and Future Work}
\label{limitations}

In this paper, we propose IMPRINT, a novel two-stage framework that achieves state-of-the-art performance in identity preservation and background harmonization for generative object compositing. We design a new pretraining scheme where the model learns a view-invariant identity-preserving representation that efficiently captures the details of the object. By decoupling the task into an identity-preserving stage and a harmonization stage, IMPRINT can generate large color and geometry variations to better align with the background. Through visual and numerical comparison results, we show that \algname{} significantly outperforms the previous methods in this task. Furthermore, we add shape guidance as an additional user control. Although \algname{} effectively addresses both identity preservation and background alignment, it has several limitations. When the required view change is too large, there could be a notable drop in identity preservation, which can be improved by exploring and incorporating a 3D model or NERF representation into our model. Another limitation is that the model may degrade consistency of small texts or logos. Potential ideas to improve this is to employ more accurate latent auto-encoder to avoid loss of information in the latent space and learn object encoders at higher resolution to encode small local details more accurately.

{
    \small
    \bibliographystyle{ieeenat_fullname}
    \bibliography{main}
}

\clearpage
\setcounter{page}{1}
\setcounter{section}{0}
\maketitlesupplementary

\section{Overview}
\label{sec:overview}

The following sections will be discussed to further support our paper:
\begin{itemize}
    \item Mask types (used for shape-guided generation);
    \item Ablation study on two alternative architectures;
    \item Additional results of shape-guided generation;
    \item Additional qualitative comparison results.
    \item Additional comparisons with AnyDoor \cite{chen2023anydoor};
    \item Failure cases;
\end{itemize}


\section{Mask Types}
\label{sec:mask_types}

\begin{figure*}[ht]
  \centering
  \begin{subfigure}{\linewidth}
  \includegraphics[width=\textwidth]{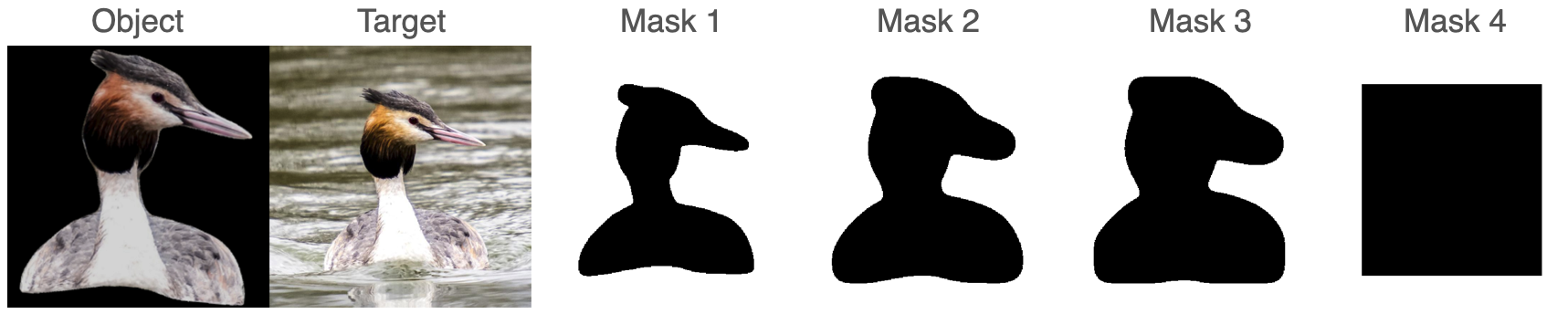}
  \end{subfigure}
  \caption{The four types of mask used in the second compositing stage. The generation is constrained in the masked area so the user-provided mask is able to modify the pose, view and shape of the subject.}
  \label{fig:mask_types}
\end{figure*}

As discussed in \cref{method_stage2}, to enable more user control, we define four levels of coarse masks, including the bounding box mask. \cref{fig:mask_types} shows all the mask types. As the coarse level increases (from mask 1 to mask 4), the model has more freedom to generate the object.

\section{Ablation Study on Alternative Architectures}
\label{sec:other_arch}

\begin{figure}
\centering
\begin{subfigure}[][][c]{0.49\textwidth}
    \includegraphics[width=\textwidth]{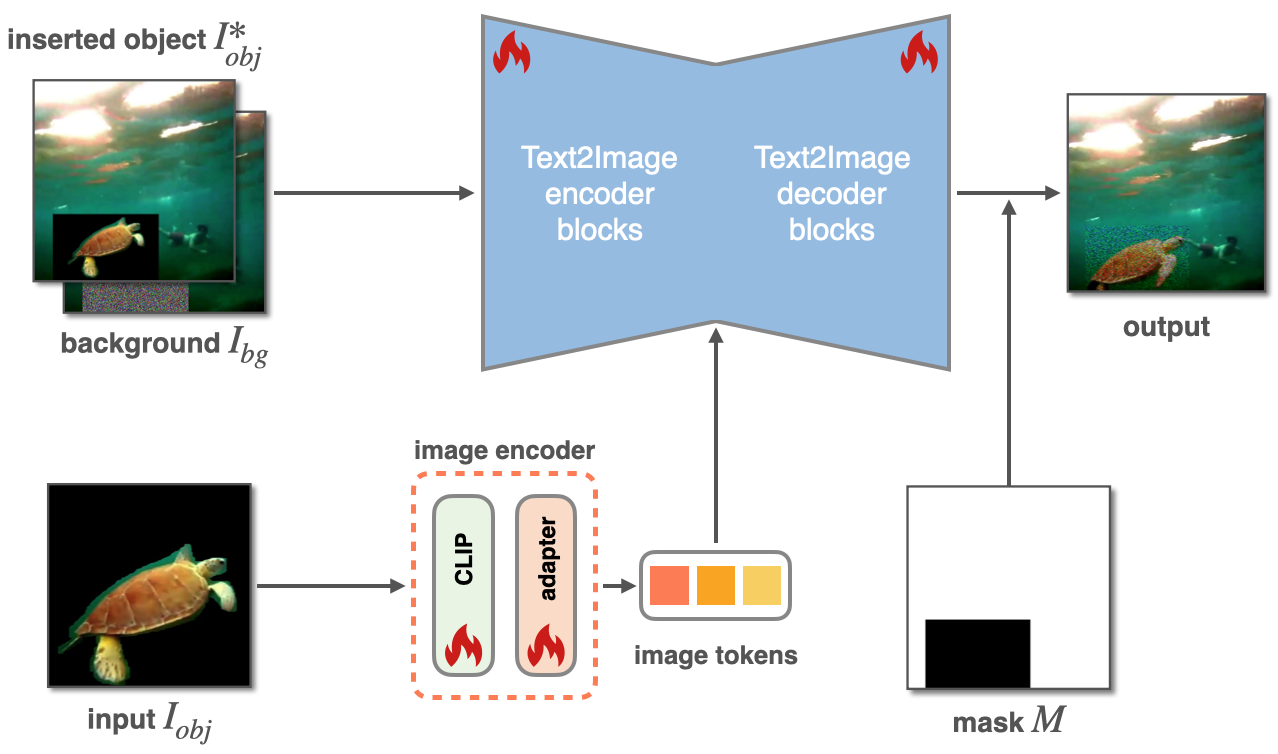}
    \caption{The concatenation-based pipeline. Aside from the embedding branch, an additional input (the inserted object $I_{obj}^*$) is concatenated with $I_{bg}$. Note that the UNet backbone encoder has 8 input channels, where the extra 4 channels are initialized as 0.0.}
    \label{fig:concat}
\end{subfigure}
\begin{subfigure}[][][c]{0.49\textwidth}
    \includegraphics[width=\textwidth]{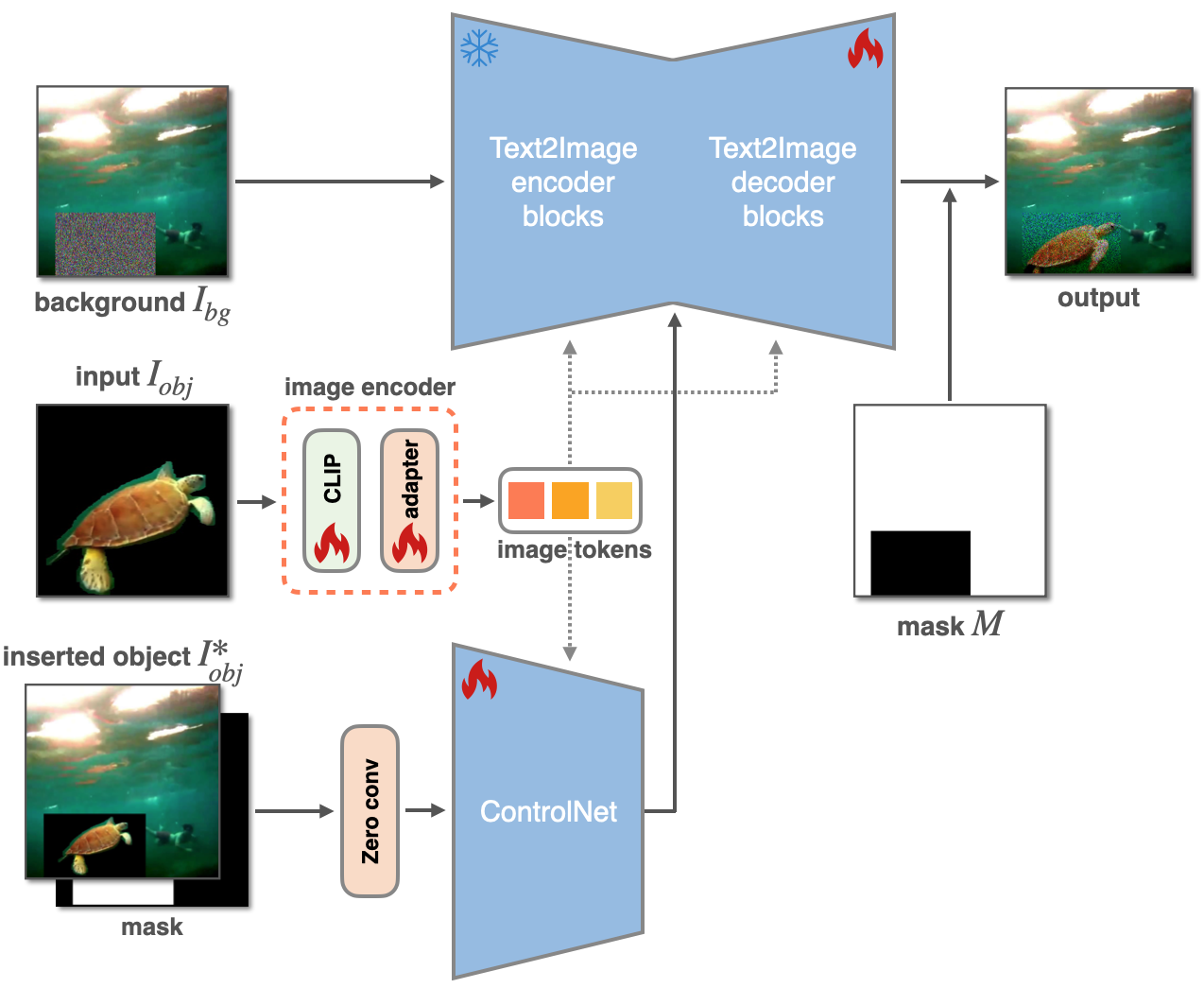}
    \caption{The ControlNet-based pipeline. In the new ControlNet branch, the concatenation of $I_{obj}^*$ and a mask is given as the additional input.}
    \label{fig:controlnet}
\end{subfigure}
\caption{The pipelines of the two alternative architectures for feature injection: Concatenation and ControlNet.}
\label{fig:other_architectures}
\end{figure}

When making efforts for better identity preservation, we also explore two alternative architectures (\cref{fig:other_architectures}) that are more intuitive to inject object features (due to the page limitation, they are removed from the main paper): 1) concatenation and 2) ControlNet \cite{zhang2023adding}. To provide extra features in this two pipelines, a naive idea is to use the same segmented object $I_{obj}$ as the additional input. However, both the structures of concatenation and ControlNet will result in a spatial correspondence between the output and the additional input (\ie, the generated object tends to have the same size and position as the input), and using $I_{obj}$ which is much larger than the mask $M$ destroys such correspondence. For this reason, we use $ I_{obj}^* $, the \textit{inserted object} image as the additional hint to provide extra features, where the cropped and resized object $I_{obj}$ is fitted in the mask area of the background image $I_{bg}$. To replace the text encoder branch, we use a combination of a CLIP encoder (ViT-L/14) and an adapter as the image encoder, fine-tuned together with the UNet backbone following the sequential collaborative training strategy discussed in \cref{method_training}. Furthermore, the two pipelines are trained on the same datasets (Pixabay and the video datasets) as our proposed model in the second stage.

\subsection{Concatenation}
\label{sec:concat}

The first architecture is illustrated in \cref{fig:concat}. An additional feature injection branch is added for the purpose of better identity preservation: $I_{obj}^*$ is concatenated with the background image $I_{bg}$. After this modification, the UNet encoder has 8 channels, where the extra 4 channels are initialized as 0.0 at the start of the training.

\subsection{ControlNet}
\label{sec:controlnet}

The second architecture is illustrated in \cref{fig:controlnet}. ControlNet is another structure to enhance spatial conditioning control, such as depth maps, Canny edges, sketches and human poses. In this pipeline, the extra inputs are fed into a trainable copy of the original UNet encoder to learn the condition. In our task of generative object compositing, we use the concatenation of the inserted object $I_{obj}^*$ and a mask $1-M$ indicating the area to generate the object.

\subsection{Quantitative Comparison}
\label{sec:quan_concat_control}

To quantize the effects of these two architectures, an evaluation is conducted on the DreamBooth dataset, just as in \cref{exp_quan}. \cref{tab:quan_concat_control} shows the results, where "Baseline" is setting 3 in the ablation study of the main paper (\cref{exp_ablation}). Our model outperforms the rest pipelines in all three metrics that measure identity preservation, demonstrating the effectiveness of IMPRINT in memorizing object details.

To further assess the compositing effects, we perform another user study with the same configuration as in the main paper (\cref{exp_user}), comparing the realism and fidelity of our results against the concatenation pipeline and ControlNet pipeline. \cref{tab:user_study_supp} displays the user preferences for different frameworks in the two questions. The results validates the superiority of our model in both ID-preserving and compositing.

\subsection{Qualitative Comparison}
\label{sec:user_study_concat_control}

\begin{figure*}
  \centering
  \begin{subfigure}{\linewidth}
  \includegraphics[width=\textwidth]{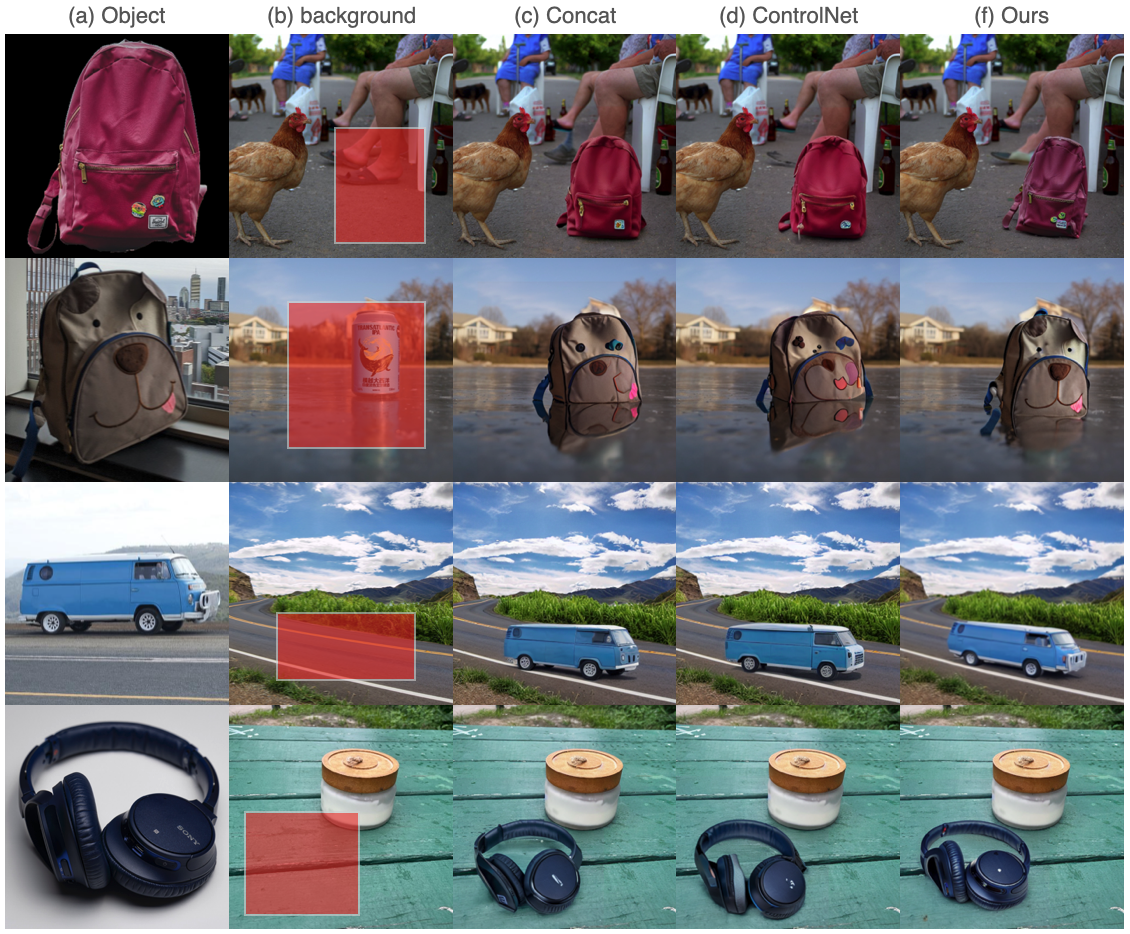}
  \end{subfigure}
  \caption{Qualitative comparisons with concatenation-based pipeline and ControlNet-based pipeline. Our model shows stronger ability in geometric adjustments (especially in the first three examples) as well as better performance in identity preservation.}
  \label{fig:qual_concat_ctrl}
\end{figure*}

\cref{fig:qual_concat_ctrl} provides a qualitative comparison between our model and the other two pipelines. Although the nature of structural correspondence in these two pipelines enhances ID preservation, it also constrains their ability to make spatial adjustments. Thus, in the figure their compositing effects are worse than our model (in the first three examples, our outputs have larger pose changes). Moreover, owing to the pretraining stage, our model achieves better performance in keeping details.

\begin{table}
\centering
\begin{adjustbox}{width=0.47\textwidth}
\begin{tabular}{lcccc}
\toprule
\textbf{Method} & \textbf{CLIP-score$\uparrow$} & \textbf{DINO-score$\uparrow$} & \textbf{DreamSim $\downarrow$} \\ 
\cmidrule{1-4}
Baseline   & 76.6250 & 39.7837 & 0.3073 \\
\cmidrule{1-4}
Concat     & 76.8125 & 40.3884 & 0.2945 \\ 
\cmidrule{1-4}
ControlNet & 76.8750 & 40.1471 & 0.2984 \\ 
\cmidrule{1-4}
Ours       & \textbf{77.0625} & \textbf{43.4463} & \textbf{0.2898} \\ 
\bottomrule
\end{tabular}
\end{adjustbox}
\caption{Quantitative comparison on the DreamBooth test set. \textit{Baseline} refers to setting 3 in the ablation study section of the main paper. Detail preservation is measured and displayed in this table, comparing our proposed model with three different architectures.
}
\label{tab:quan_concat_control}
\end{table}

\begin{table}
  \small
  \centering
  \begin{tabular}{l|ll|ll}
    \hlineB{2}
     & Ours & Concat & Ours & ControlNet \\ \hline
    Realism  & \textbf{50.68} & 49.32 & \textbf{53.38} & 46.62 \\ 
    Fidelity & \textbf{55.41} & 44.59 & \textbf{54.73} & 45.27 \\ \hlineB{2}
  \end{tabular}
  \caption{User study results (in percentage). In the two questions that evaluates reality and similarity, the workers are presented with side-by-side results from different models and are asked to make comparison.}
  \label{tab:user_study_supp}
\end{table}

\section{Additional Results of Shape-Guided Generation}
\label{sec:mask_ctrl_supp}

\subsection{Ablation Study}

Shape-guidance is an important feature supported by our model that enables more user control. This feature is not independent of our efforts in identity preservation. Instead, the overall performance (realism and fidelity) of shape-guided generation is improved by our pretraining stage, as demonstrated by \cref{tab:pre_on_mask_ctrl}.

This ablation study is conducted on the video datasets (the test sets). We follow the same data generation pipeline in \cref{method_data}: the target image and the input object are taken from frames $I_{n1}, I_{n2}$ respectively, with $n1 \neq n2$. The guidance mask $M$ is a coarse mask of the object segmentation in the target frame $n1$. We compare our proposed model with another model that is only trained on the second compositing stage. The quantitative results show the improvement of the pretraining stage.

\begin{table}
\centering
\begin{adjustbox}{width=0.47\textwidth}
\begin{tabular}{lccccc}
\toprule
\textbf{Method} & \textbf{FID $\downarrow$} & \textbf{CLIP-score$\uparrow$} & \textbf{DINO-score$\uparrow$} & \textbf{DreamSim $\downarrow$} \\ 
\cmidrule{1-5}
No PRE & 70.0528 & 91.5625 & 83.8687 & 0.1723 \\
\cmidrule{1-5}
PRE    & \textbf{59.6255} & \textbf{91.9375} & \textbf{84.7372} & \textbf{0.1589} \\ 
\bottomrule
\end{tabular}
\end{adjustbox}
\caption{Ablation study on the pretraining stage in shape-guided generation. \textit{PRE} means the pretraining. When the pretraining is finished, the model shows stronger capabilities in ID-preserving and realism, highlighting the fact that our pretraining boosts the performance of shape-guided generation.
}
\label{tab:pre_on_mask_ctrl}
\end{table}


\section{Additional Qualitative Results}
\label{sec:qual_comp_supp}

\begin{figure*}
  \centering
  \begin{subfigure}{\linewidth}
  \includegraphics[width=\textwidth]{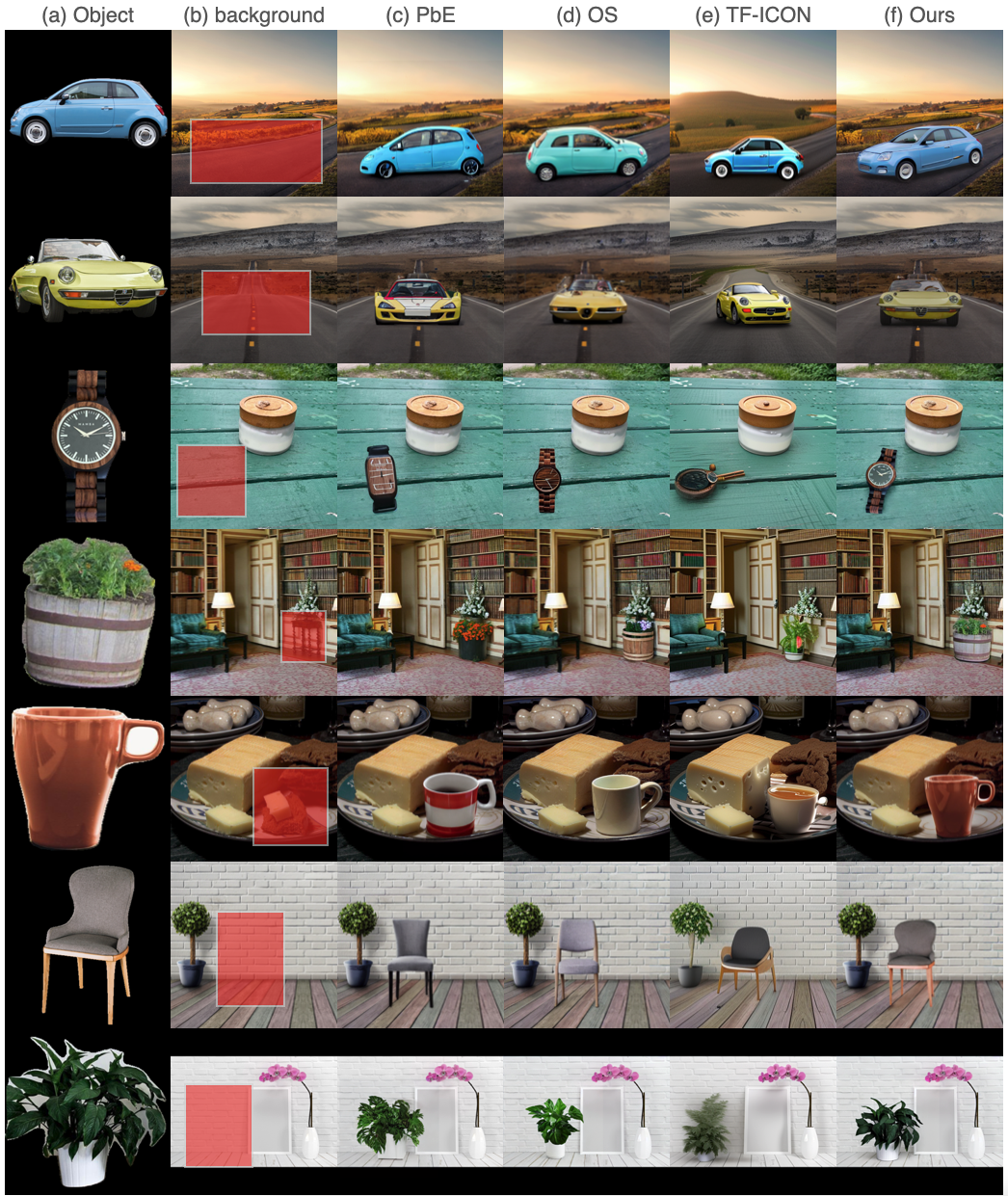}
  \end{subfigure}
  \caption{More qualitative comparisons. We compare our proposed model with Paint-by-Example (PbE), ObjectStitch (OS) and TF-ICON. IMPRINT better preserves object identity and the generated object is more consistent with the background.}
  \label{fig:qual_supp1}
\end{figure*}

\begin{figure*}
  \centering
  \begin{subfigure}{\linewidth}
  \includegraphics[width=\textwidth]{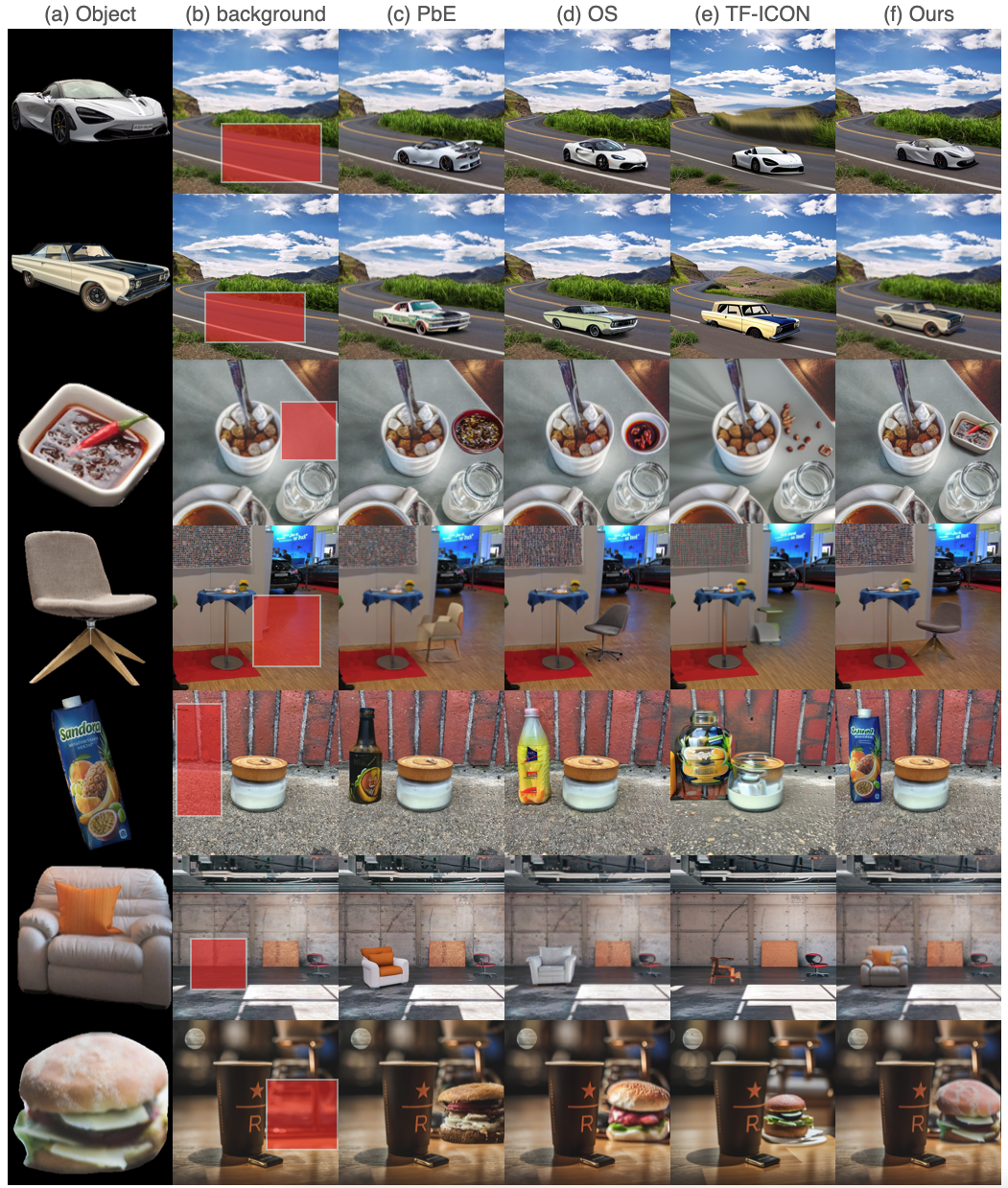}
  \end{subfigure}
  \caption{More qualitative comparisons. We compare our proposed model with Paint-by-Example (PbE), ObjectStitch (OS) and TF-ICON. IMPRINT better preserves object identity and the generated object is more consistent with the background.}
  \label{fig:qual_supp2}
\end{figure*}

To further show the advantages of our model against the baseline methods (Paint-by-Example or PbE \cite{yang2023paint}, ObjectStitch or OS \cite{song2023objectstitch} and TF-ICON \cite{lu2023tf}), we include more qualitative results in \cref{fig:qual_supp1} and \cref{fig:qual_supp2}.

\section{Additional Comparisons with AnyDoor}
\label{sec:comp_anydoor}

\begin{table}[h]
    \centering
    \begin{tabular}{c c}
        \begin{adjustbox}{width=0.21\textwidth}
        \begin{tabular}{c|cc}
            \hlineB{2}
            \textbf{Method} & \textbf{CLIP$\uparrow$} & \textbf{DINO$\uparrow$} \\
            \hline
            AnyDoor & 83.563 & 83.598 \\
            \hline
            Ours & \textbf{85.813} & \textbf{86.589} \\ \hlineB{2}
        \end{tabular}
        \end{adjustbox}
        &
        \hspace{-12pt}
        \begin{adjustbox}{width=0.22\textwidth}
        \begin{tabular}{c|cc}
            \hlineB{2}
            \textbf{Method} & \textbf{Realism} & \textbf{Fidelity} \\
            \hline
            AnyDoor & 40.71 & 35.18 \\
            \hline
            Ours & \textbf{59.29} & \textbf{64.82} \\ \hlineB{2}
        \end{tabular}
        \end{adjustbox}
    \end{tabular}
    \caption{
    Left: Quantitative comparison on the DreamBooth test set.
    Right: User study results (in percentage).
    }
\label{tab:add_quan_anydoor}
\end{table}

\begin{figure*}
  \centering
  \begin{subfigure}{\linewidth}
  \includegraphics[width=\textwidth]{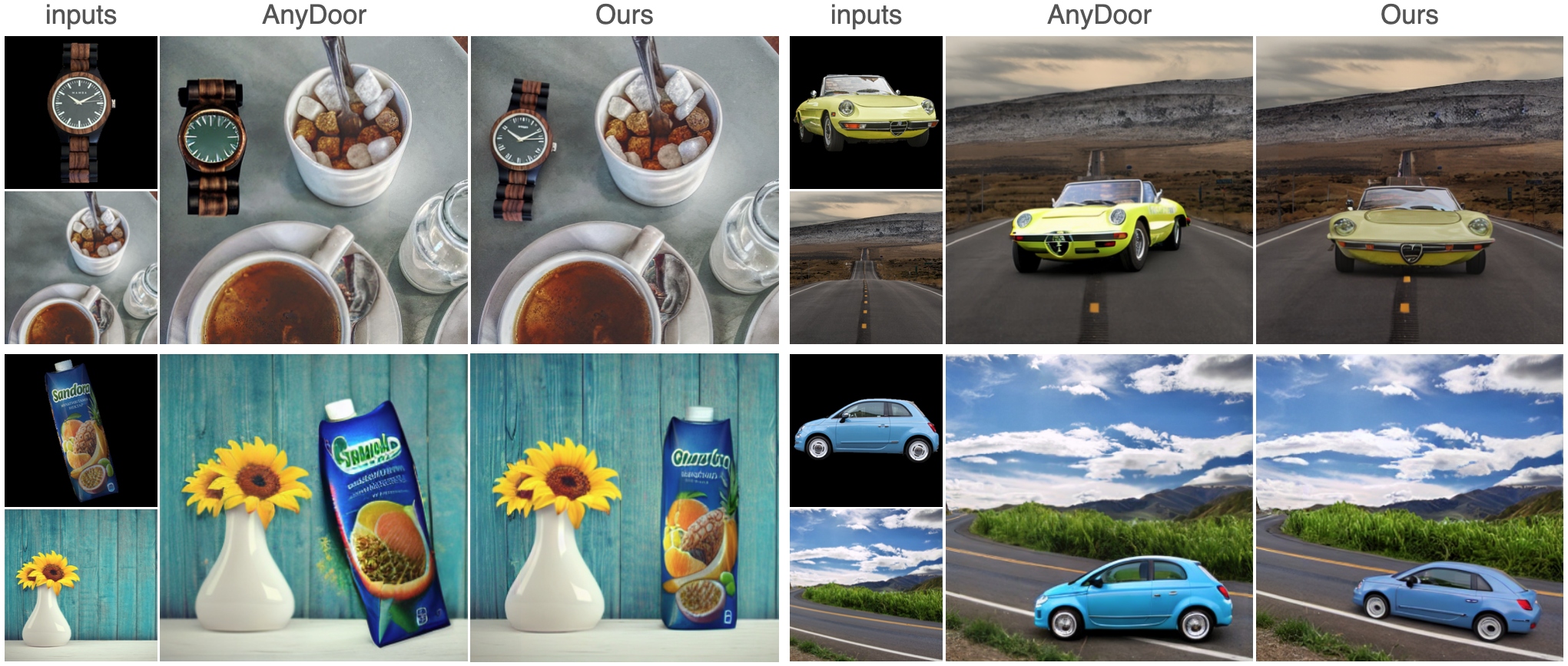}
  \end{subfigure}
  \caption{Additional qualitative comparisons with AnyDoor.}
  \label{fig:add_qual_anydoor}
\end{figure*}

We provide additional comparisons below using the official implementation of AnyDoor. We observe that IMPRINT significantly outperforms AnyDoor in the following experiments:

\begin{itemize}
    \item We calculate CLIP score and DINO score on the DreamBooth test set to measure the identity preservation (as shown in the left of \cref{tab:add_quan_anydoor}). Note that to get more accurate results, we masked the background of all generated images when performing the evaluation on the DreamBooth set.
    \item We conduct a new user study under the same setting as the user study in the main paper (shown in the right of \cref{tab:add_quan_anydoor}). The users have higher preference rate in our results in both realism and detail preservation.
    \item In the additional visual comparisons in \cref{fig:add_qual_anydoor}, our model demonstrates greater adaptability in adjusting the object's pose to match the background, while preserving the details.
\end{itemize}


\section{Failure Cases}
\label{sec:failure}

\begin{figure*}
  \centering
  \begin{subfigure}{\linewidth}
  \includegraphics[width=\textwidth]{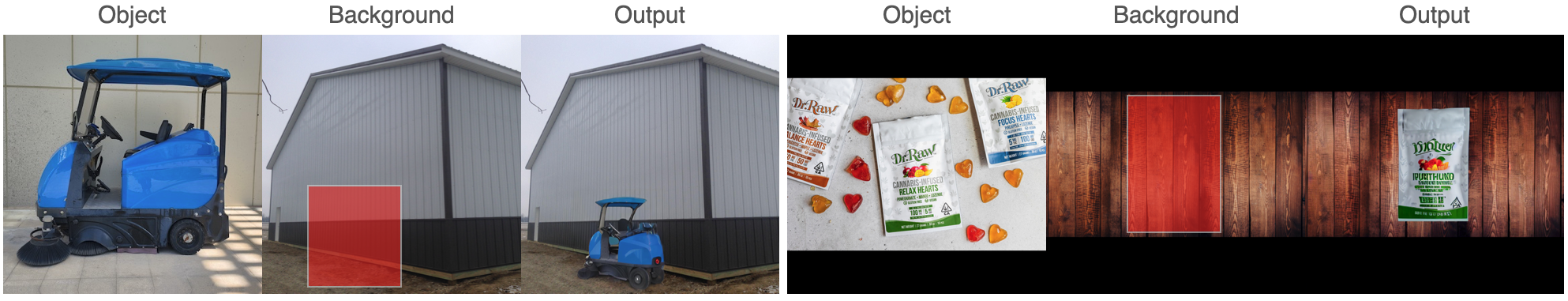}
  \end{subfigure}
  \caption{Limitations. 1) The first example shows identity loss when making large geometric corrections. The structure of the vehicle changes after generation. 2) The second example shows the degradation of small logos and texts after decoding from the latent space.
  }
  \label{fig:limit}
\end{figure*}

\cref{fig:limit} shows the limitations of IMPRINT, as discussed in \cref{limitations}. In the first example, Though the vehicle is well aligned with the background, its structure is deformed and partially lose its identity due to the large spatial transformation. In the second example, the small logos and texts on the item cannot be fully maintained and exhibits small artifacts, mainly caused by the decoder in Stable Diffusion \cite{rombach2022high}.

\end{document}